\documentclass[letterpaper, 10 pt, journal, twoside]{IEEEtran}

\IEEEoverridecommandlockouts                              

\usepackage{graphicx} 
\usepackage{amsmath} 
\usepackage{amssymb}  
\usepackage{makecell}
\usepackage{physics}
\usepackage[ruled, vlined, linesnumbered, nofillcomment]{algorithm2e}
\usepackage{array}
\usepackage{tabularx}
\usepackage{adjustbox}
\usepackage{makecell}
\usepackage{multicol}
\usepackage{bbm}
\usepackage{multirow}
\usepackage[bookmarks=true,hidelinks]{hyperref} 
\hypersetup{%
  colorlinks=true,%
  linkcolor=blue,
  citecolor=blue,
  urlcolor=black,
}
\usepackage[caption=false]{subfig}
\usepackage{booktabs}
\usepackage{comment}
\usepackage{bm}
\usepackage{threeparttable}
\usepackage{cleveref}
\usepackage[table,xcdraw,dvipsnames]{xcolor}
\usepackage{soul} 
\usepackage{hhline}
\usepackage{arydshln} 
\usepackage{cancel}
\usepackage{graphbox} 
\usepackage{tikz}
\usepackage{listofitems} 
\usetikzlibrary{calc}
\usepackage{pgfplots}  

\usepackage{datamanager} 

\DeclareMathAlphabet{\mathpzc}{OT1}{pzc}{m}{it}

\DeclareMathOperator*{\argmin}{arg\,min}
\DeclareMathOperator{\diag}{diag}

\newcommand{\mat}{\bm}

\newcommand{\eg}{{\em e.g.}}
\newcommand{\ie}{{\em i.e.}}

\newcommand{\dscomment}[1]{{\color{red} \hl{#1}}}

\newcommand{\maybevspace}[1]{\dscomment{Remove \textbackslash maybevspace}}

\usepackage{cellspace}
\usepackage{scalerel}
\usetikzlibrary{svg.path}
\definecolor{orcidlogocol}{HTML}{A6CE39}
\tikzset{
    orcidlogo/.pic={
        \fill[orcidlogocol] svg{M256,128c0,70.7-57.3,128-128,128C57.3,256,0,198.7,0,128C0,57.3,57.3,0,128,0C198.7,0,256,57.3,256,128z};
        \fill[white] svg{M86.3,186.2H70.9V79.1h15.4v48.4V186.2z}
        svg{M108.9,79.1h41.6c39.6,0,57,28.3,57,53.6c0,27.5-21.5,53.6-56.8,53.6h-41.8V79.1z M124.3,172.4h24.5c34.9,0,42.9-26.5,42.9-39.7c0-21.5-13.7-39.7-43.7-39.7h-23.7V172.4z}
        svg{M88.7,56.8c0,5.5-4.5,10.1-10.1,10.1c-5.6,0-10.1-4.6-10.1-10.1c0-5.6,4.5-10.1,10.1-10.1C84.2,46.7,88.7,51.3,88.7,56.8z};
    }
}

\DeclareFontFamily{OT1}{pzc}{}
\DeclareFontShape{OT1}{pzc}{m}{it}{<-> s * [1.05] pzcmi7t}{}
\DeclareMathAlphabet{\mathpzc}{OT1}{pzc}{m}{it}

\newcommand\orcidicon[1]{\href{https://orcid.org/#1}{\mbox{\scalerel*{
                \begin{tikzpicture}[yscale=-1,transform shape]
                \pic{orcidlogo};
                \end{tikzpicture}
            }{|}}}}

\colorlet{colorFst}{Green!25}       
\colorlet{colorSnd}{SpringGreen!35}  
\colorlet{colorTrd}{Yellow!15}      
\colorlet{colorLow}{darkgray!30}    
\newcommand{\markfst}[1]{\textbf{#1}}   
\newcommand{\marksnd}[1]{\underline{#1}}      

\dmrecord{c3g}{
  depth-error-inferno.room-0-2-2 = 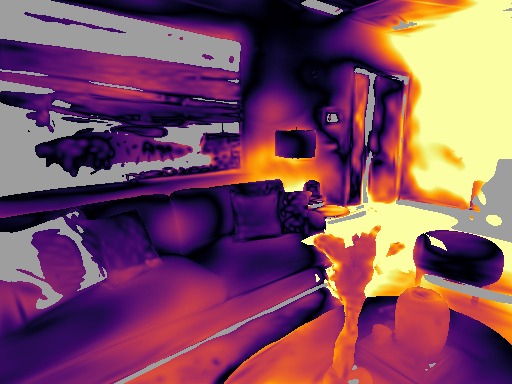,
  depth-error-inferno.room-0-2-4 = 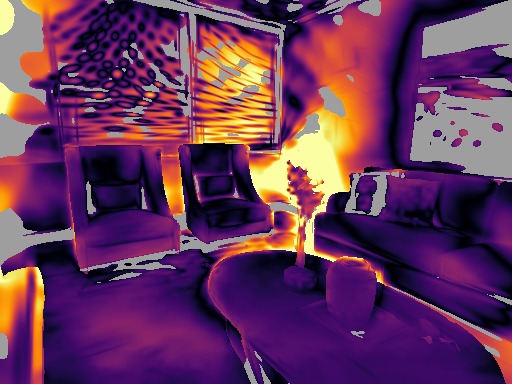,
  depth-turbo.room-0-2-2         = 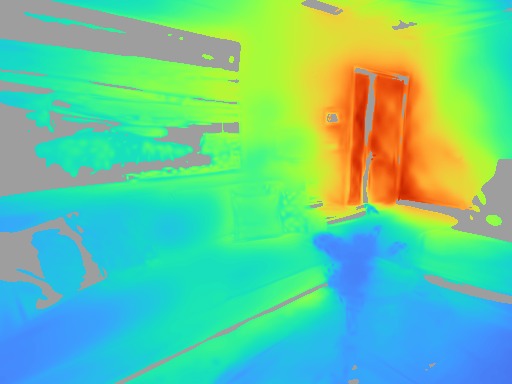,
  depth-turbo.room-0-2-4         = 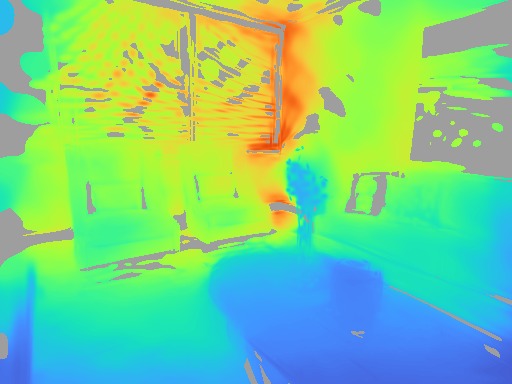,
  rep2.abs-diff                  = 0.271,
  rep2.abs-rel                   = 12.8,
  rep2.coverage                  = 82.6,
  rep2.delta-125                 = 85.6,
  rep2.fps                       = 17.1,
  rep2.lpips                     = 0.379,
  rep2.peak-inf-mb               = 264.4,
  rep2.peak-inf-mb@fmt           = \marksnd,
  rep2.psnr                      = 19.57,
  rep2.ssim                      = 0.724,
  rep2.total-mem-mb              = 4129.3,
  rep2.weights-mb                = 3864.9,
  rgb.room-0-2-2                 = 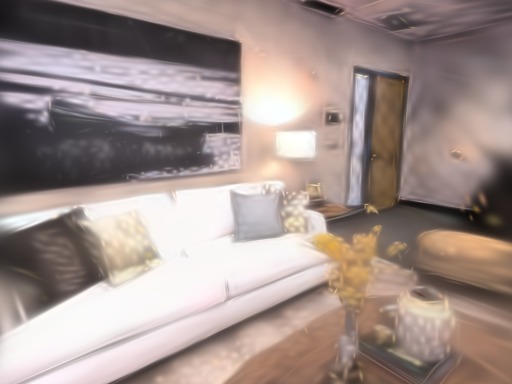,
  rgb.room-0-2-4                 = 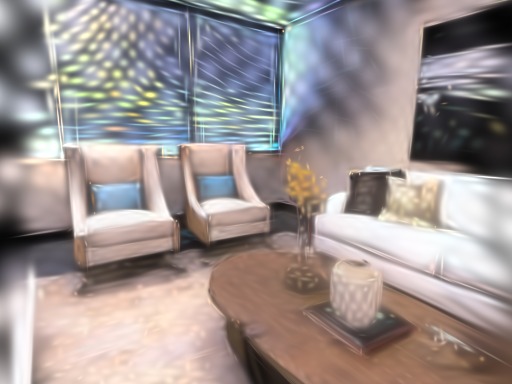,
}

\dmrecord{freesplat}{
  depth-error-inferno.room-0-2-2 = 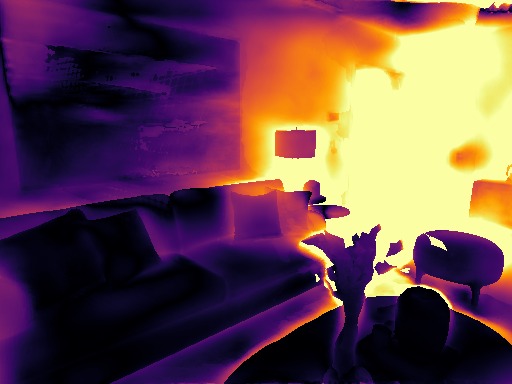,
  depth-error-inferno.room-0-2-4 = 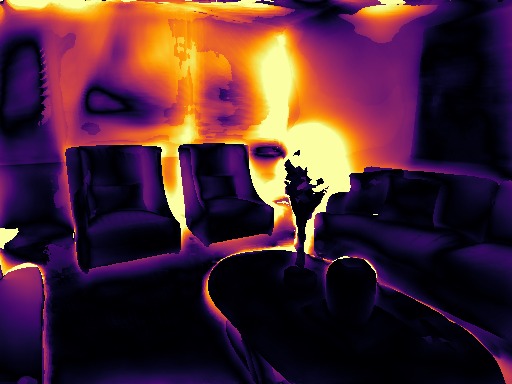,
  depth-turbo.room-0-2-2         = 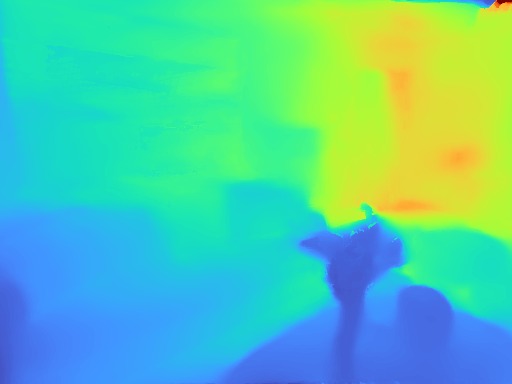,
  depth-turbo.room-0-2-4         = 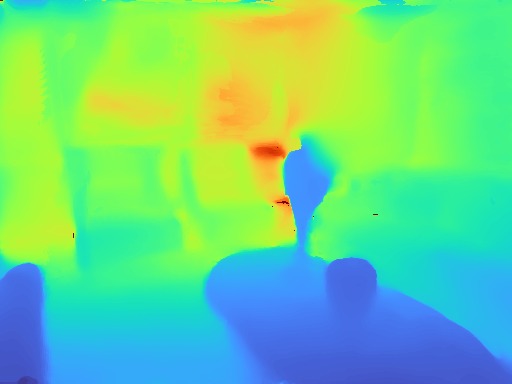,
  rep2.abs-diff                  = 0.216,
  rep2.abs-diff@fmt              = \marksnd,
  rep2.abs-rel                   = 9.0,
  rep2.abs-rel@fmt               = \marksnd,
  rep2.coverage                  = 96.6,
  rep2.coverage@fmt              = \markfst,
  rep2.delta-125                 = 90.6,
  rep2.delta-125@fmt             = \marksnd,
  rep2.fps                       = 12.8,
  rep2.lpips                     = 0.142,
  rep2.lpips@fmt                 = \marksnd,
  rep2.peak-inf-mb               = 985.6,
  rep2.psnr                      = 30.12,
  rep2.psnr@fmt                  = \markfst,
  rep2.ssim                      = 0.900,
  rep2.ssim@fmt                  = \markfst,
  rep2.total-mem-mb              = 1179.1,
  rep2.weights-mb                = 193.5,
  rgb.room-0-2-2                 = 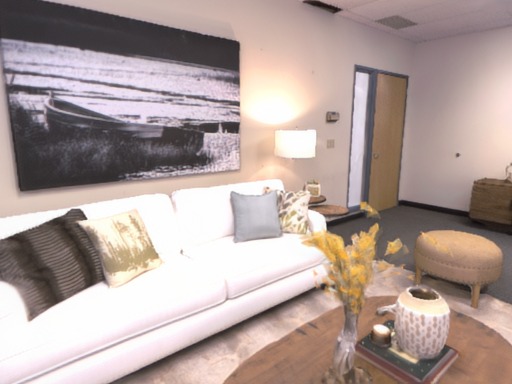,
  rgb.room-0-2-4                 = 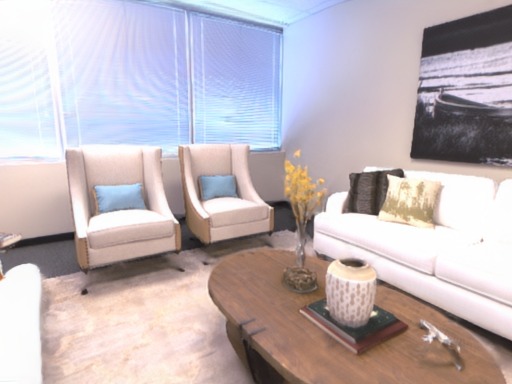,
}

\dmrecord{g2ba}{
  depth-error-inferno.room-0-2-2 = 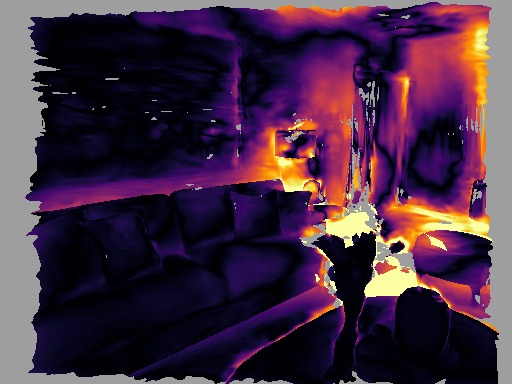,
  depth-error-inferno.room-0-2-4 = 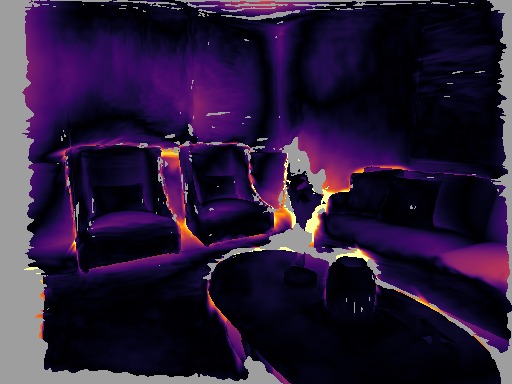,
  depth-turbo.room-0-2-2         = 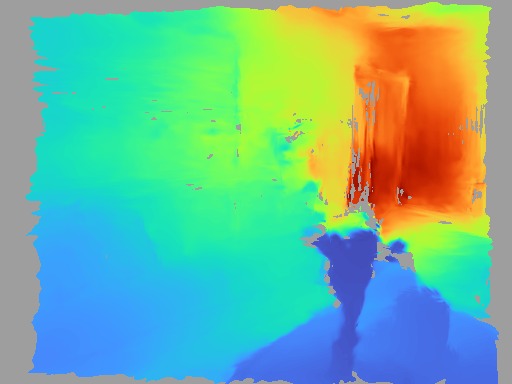,
  depth-turbo.room-0-2-4         = 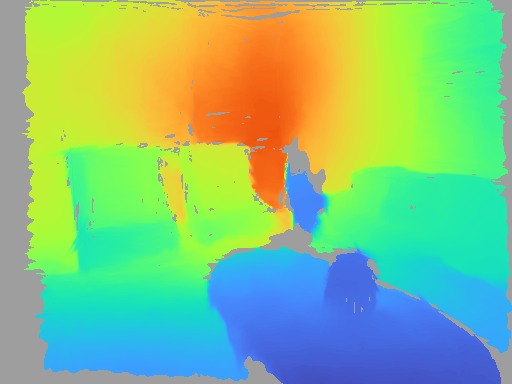,
  rep2.abs-diff                  = 0.091,
  rep2.abs-diff@fmt              = \markfst,
  rep2.abs-rel                   = 4.0,
  rep2.abs-rel@fmt               = \markfst,
  rep2.coverage                  = 77.6,
  rep2.delta-125                 = 98.1,
  rep2.delta-125@fmt             = \markfst,
  rep2.fps                       = 89.4,
  rep2.fps@fmt                   = \markfst,
  rep2.lpips                     = 0.446,
  rep2.peak-inf-mb               = 88.1,
  rep2.peak-inf-mb@fmt           = \markfst,
  rep2.psnr                      = 23.36,
  rep2.ssim                      = 0.739,
  rep2.total-mem-mb              = 115.4,
  rep2.total-mem-mb@fmt          = \markfst,
  rep2.weights-mb                = 27.3,
  rep2.weights-mb@fmt            = \markfst,
  rgb.room-0-2-2                 = 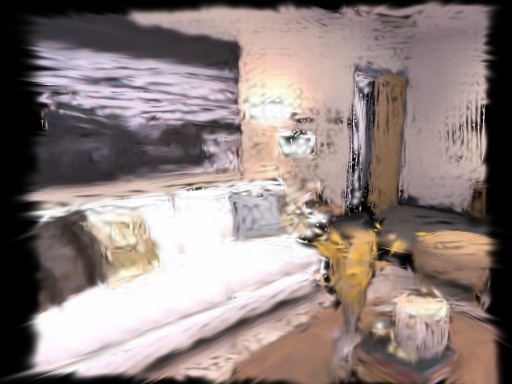,
  rgb.room-0-2-4                 = 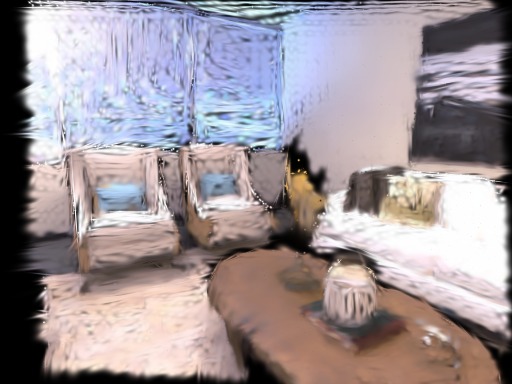,
}

\dmrecord{gt}{
  depth-turbo.room-0-2-2 = 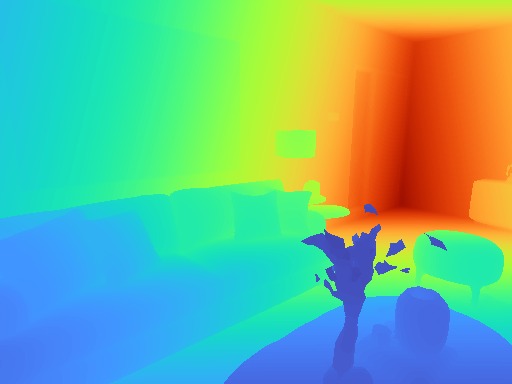,
  depth-turbo.room-0-2-4 = 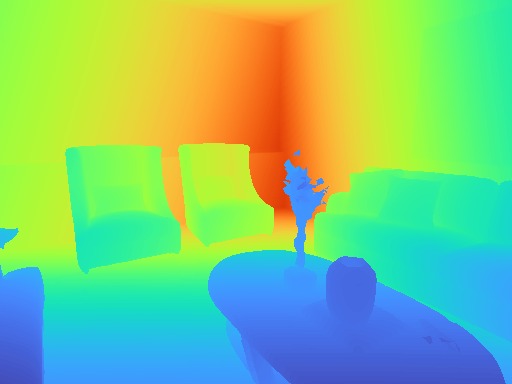,
  rgb.room-0-2-2         = 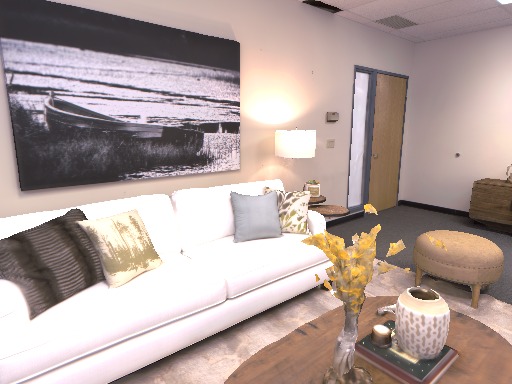,
  rgb.room-0-2-4         = 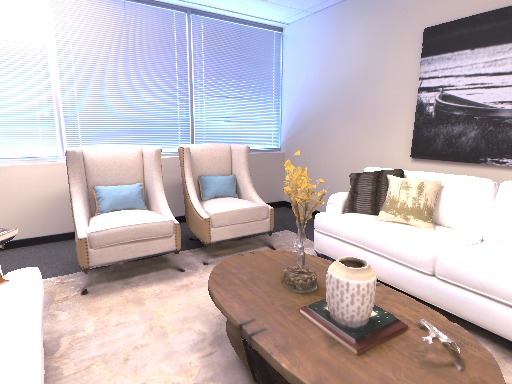,
}

\dmrecord{monosplat}{
  depth-error-inferno.room-0-2-2 = 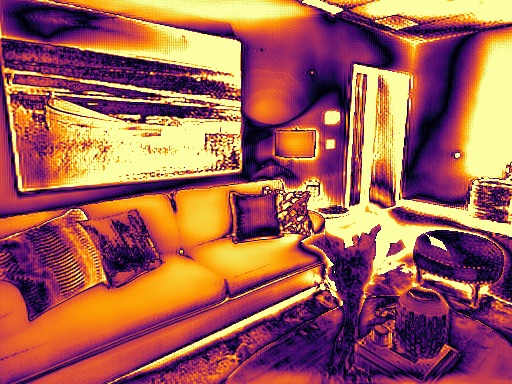,
  depth-error-inferno.room-0-2-4 = 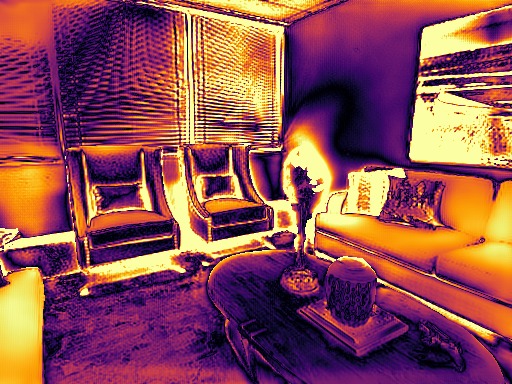,
  depth-turbo.room-0-2-2         = 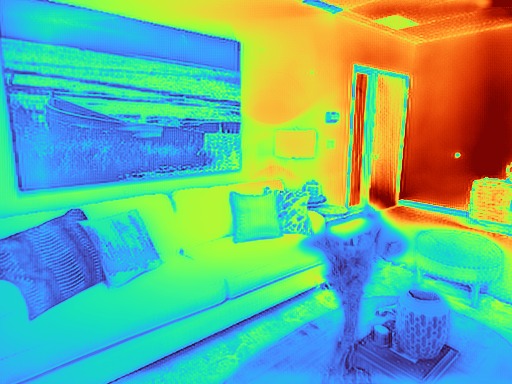,
  depth-turbo.room-0-2-4         = 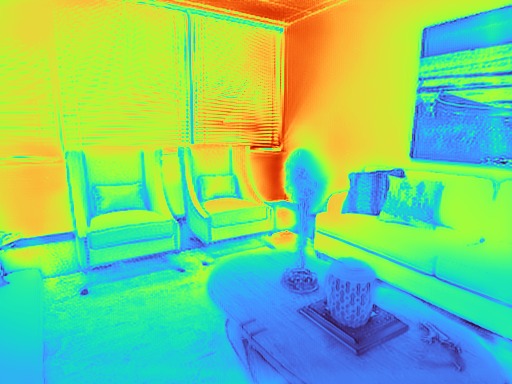,
  rep2.abs-diff                  = 0.504,
  rep2.abs-rel                   = 22.4,
  rep2.coverage                  = 96.6,
  rep2.coverage@fmt              = \marksnd,
  rep2.delta-125                 = 62.5,
  rep2.fps                       = 37.3,
  rep2.fps@fmt                   = \marksnd,
  rep2.lpips                     = 0.180,
  rep2.peak-inf-mb               = 610.9,
  rep2.psnr                      = 22.24,
  rep2.ssim                      = 0.866,
  rep2.ssim@fmt                  = \marksnd,
  rep2.total-mem-mb              = 745.4,
  rep2.total-mem-mb@fmt          = \marksnd,
  rep2.weights-mb                = 134.5,
  rgb.room-0-2-2                 = 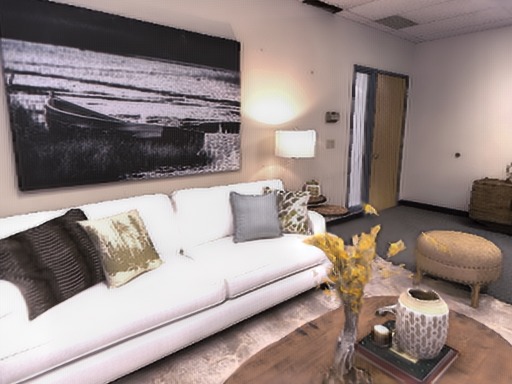,
  rgb.room-0-2-4                 = 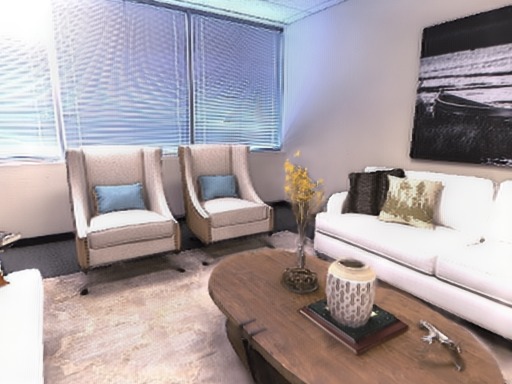,
}

\dmrecord{mvsplat}{
  depth-error-inferno.room-0-2-2 = 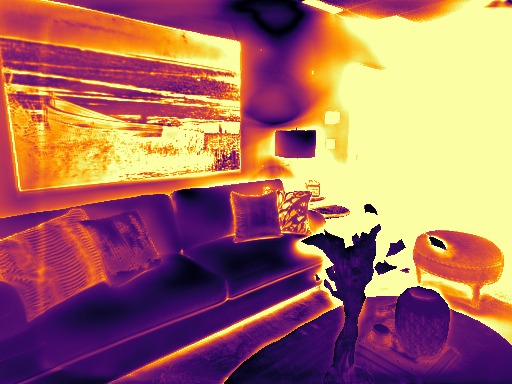,
  depth-error-inferno.room-0-2-4 = 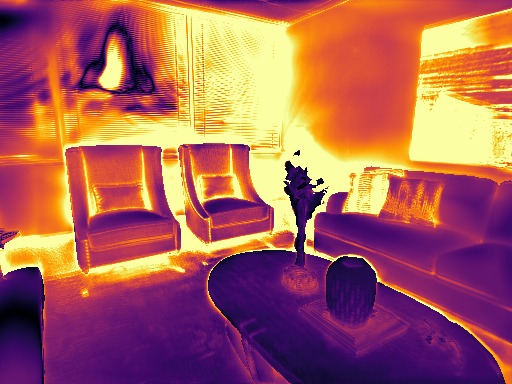,
  depth-turbo.room-0-2-2         = 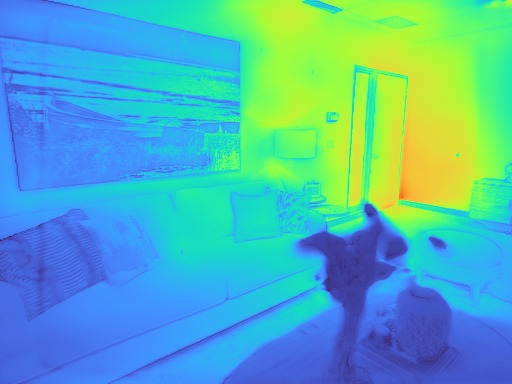,
  depth-turbo.room-0-2-4         = 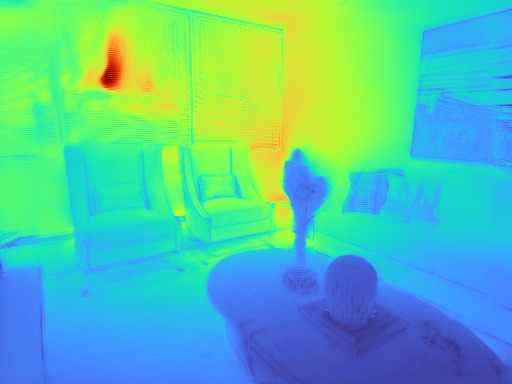,
  rep2.abs-diff                  = 0.609,
  rep2.abs-rel                   = 24.1,
  rep2.coverage                  = 95.9,
  rep2.delta-125                 = 46.9,
  rep2.fps                       = 9.9,
  rep2.lpips                     = 0.135,
  rep2.lpips@fmt                 = \markfst,
  rep2.peak-inf-mb               = 3120.0,
  rep2.psnr                      = 27.22,
  rep2.psnr@fmt                  = \marksnd,
  rep2.ssim                      = 0.860,
  rep2.total-mem-mb              = 3165.6,
  rep2.weights-mb                = 45.6,
  rep2.weights-mb@fmt            = \marksnd,
  rgb.room-0-2-2                 = 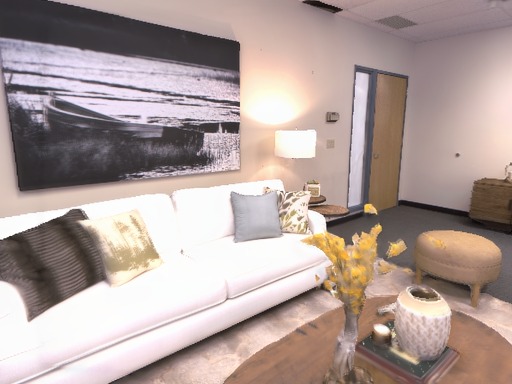,
  rgb.room-0-2-4                 = 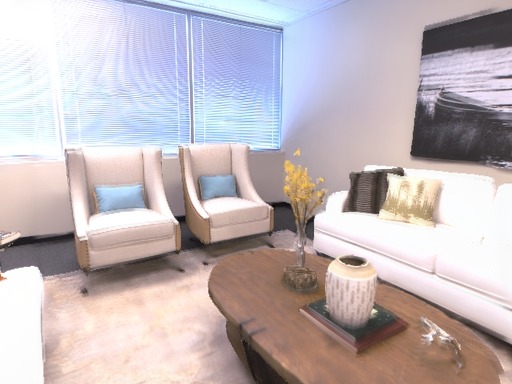,
}

\dmrecord{surfelsplat}{
  depth-error-inferno.room-0-2-2 = 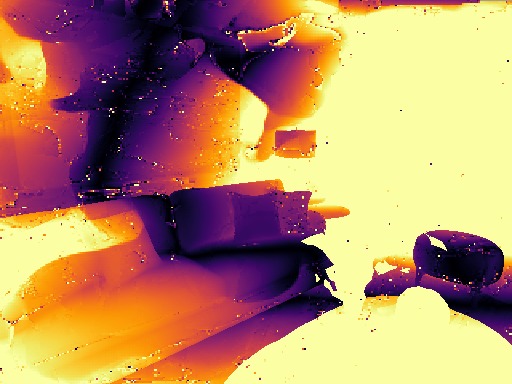,
  depth-error-inferno.room-0-2-4 = 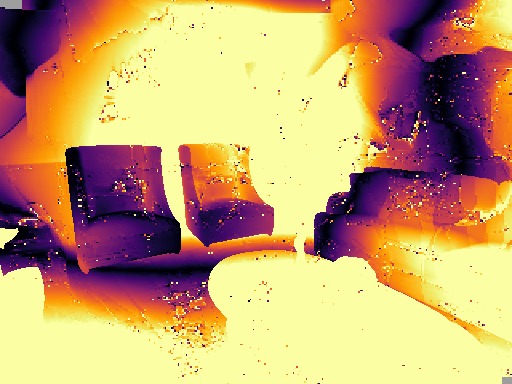,
  depth-turbo.room-0-2-2         = 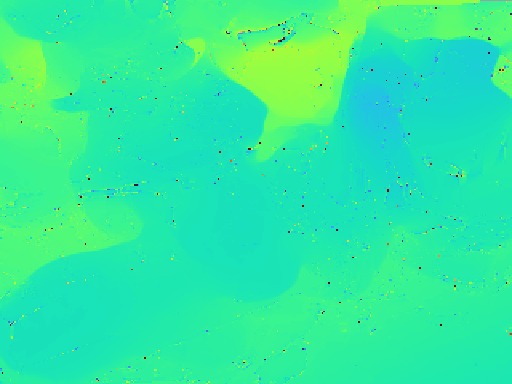,
  depth-turbo.room-0-2-4         = 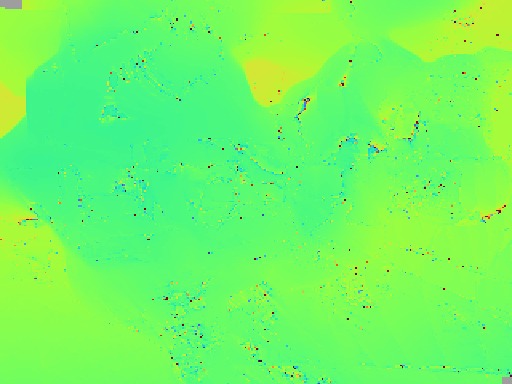,
  rep2.abs-diff                  = 0.627,
  rep2.abs-rel                   = 30.1,
  rep2.coverage                  = 96.0,
  rep2.delta-125                 = 53.7,
  rep2.fps                       = 5.0,
  rep2.lpips                     = 0.435,
  rep2.peak-inf-mb               = 10351.4,
  rep2.psnr                      = 18.13,
  rep2.ssim                      = 0.657,
  rep2.total-mem-mb              = 12385.5,
  rep2.weights-mb                = 2034.1,
  rgb.room-0-2-2                 = 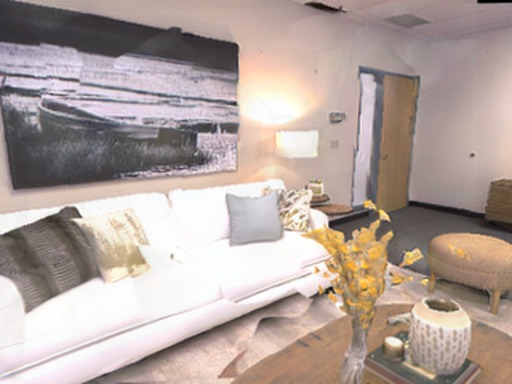,
  rgb.room-0-2-4                 = 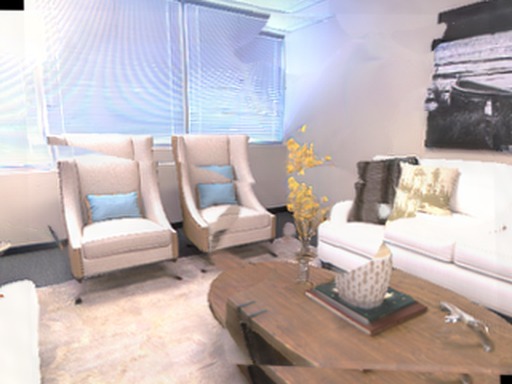,
}

\dmrecord{c3g}{
  rep3.abs-diff        = 0.223,
  rep3.abs-rel         = 10.7,
  rep3.coverage        = 90.2,
  rep3.delta-125       = 89.9,
  rep3.fps             = 14.5,
  rep3.lpips           = 0.337,
  rep3.peak-inf-mb     = 332.1,
  rep3.peak-inf-mb@fmt = \marksnd,
  rep3.psnr            = 21.80,
  rep3.ssim            = 0.773,
  rep3.total-mem-mb    = 4197.0,
  rep3.weights-mb      = 3864.9,
}

\dmrecord{freesplat}{
  rep3.abs-diff      = 0.141,
  rep3.abs-diff@fmt  = \marksnd,
  rep3.abs-rel       = 6.3,
  rep3.abs-rel@fmt   = \marksnd,
  rep3.coverage      = 97.4,
  rep3.coverage@fmt  = \markfst,
  rep3.delta-125     = 94.4,
  rep3.delta-125@fmt = \marksnd,
  rep3.fps           = 10.3,
  rep3.lpips         = 0.137,
  rep3.lpips@fmt     = \markfst,
  rep3.peak-inf-mb   = 1480.2,
  rep3.psnr          = 30.53,
  rep3.psnr@fmt      = \markfst,
  rep3.ssim          = 0.906,
  rep3.ssim@fmt      = \markfst,
  rep3.total-mem-mb  = 1673.7,
  rep3.weights-mb    = 193.5,
}

\dmrecord{g2ba}{
  rep3.abs-diff         = 0.106,
  rep3.abs-diff@fmt     = \markfst,
  rep3.abs-rel          = 4.4,
  rep3.abs-rel@fmt      = \markfst,
  rep3.coverage         = 85.8,
  rep3.delta-125        = 97.5,
  rep3.delta-125@fmt    = \markfst,
  rep3.fps              = 71.5,
  rep3.fps@fmt          = \markfst,
  rep3.lpips            = 0.417,
  rep3.peak-inf-mb      = 175.9,
  rep3.peak-inf-mb@fmt  = \markfst,
  rep3.psnr             = 24.07,
  rep3.ssim             = 0.781,
  rep3.total-mem-mb     = 203.2,
  rep3.total-mem-mb@fmt = \markfst,
  rep3.weights-mb       = 27.3,
  rep3.weights-mb@fmt   = \markfst,
}

\dmrecord{gt}{
}

\dmrecord{monosplat}{
  rep3.abs-diff         = 0.366,
  rep3.abs-rel          = 16.4,
  rep3.coverage         = 97.3,
  rep3.coverage@fmt     = \marksnd,
  rep3.delta-125        = 76.0,
  rep3.fps              = 30.0,
  rep3.fps@fmt          = \marksnd,
  rep3.lpips            = 0.166,
  rep3.lpips@fmt        = \marksnd,
  rep3.peak-inf-mb      = 916.1,
  rep3.psnr             = 27.12,
  rep3.psnr@fmt         = \marksnd,
  rep3.ssim             = 0.897,
  rep3.ssim@fmt         = \marksnd,
  rep3.total-mem-mb     = 1050.6,
  rep3.total-mem-mb@fmt = \marksnd,
  rep3.weights-mb       = 134.5,
}

\dmrecord{mvsplat}{
  rep3.abs-diff       = 0.453,
  rep3.abs-rel        = 19.7,
  rep3.coverage       = 96.6,
  rep3.delta-125      = 69.4,
  rep3.fps            = 5.3,
  rep3.lpips          = 0.194,
  rep3.peak-inf-mb    = 4878.0,
  rep3.psnr           = 25.04,
  rep3.ssim           = 0.837,
  rep3.total-mem-mb   = 4923.6,
  rep3.weights-mb     = 45.6,
  rep3.weights-mb@fmt = \marksnd,
}

\dmrecord{surfelsplat}{
  rep3.abs-diff     = 0.621,
  rep3.abs-rel      = 30.3,
  rep3.coverage     = 96.9,
  rep3.delta-125    = 53.7,
  rep3.fps          = 2.4,
  rep3.lpips        = 0.466,
  rep3.peak-inf-mb  = 15527.1,
  rep3.psnr         = 17.43,
  rep3.ssim         = 0.647,
  rep3.total-mem-mb = 17561.2,
  rep3.weights-mb   = 2034.1,
}

\dmrecord{c3g}{
  sca2.abs-diff        = 0.280,
  sca2.abs-rel         = 15.5,
  sca2.coverage        = 91.7,
  sca2.delta-125       = 78.6,
  sca2.fps             = 17.0,
  sca2.lpips           = 0.529,
  sca2.peak-inf-mb     = 264.4,
  sca2.peak-inf-mb@fmt = \marksnd,
  sca2.psnr            = 17.48,
  sca2.ssim            = 0.669,
  sca2.total-mem-mb    = 4129.3,
  sca2.weights-mb      = 3864.9,
}

\dmrecord{freesplat}{
  sca2.abs-diff      = 0.172,
  sca2.abs-diff@fmt  = \marksnd,
  sca2.abs-rel       = 9.2,
  sca2.abs-rel@fmt   = \marksnd,
  sca2.coverage      = 95.0,
  sca2.coverage@fmt  = \markfst,
  sca2.delta-125     = 89.9,
  sca2.delta-125@fmt = \marksnd,
  sca2.fps           = 12.9,
  sca2.lpips         = 0.224,
  sca2.lpips@fmt     = \markfst,
  sca2.peak-inf-mb   = 983.2,
  sca2.psnr          = 27.28,
  sca2.psnr@fmt      = \markfst,
  sca2.ssim          = 0.829,
  sca2.ssim@fmt      = \markfst,
  sca2.total-mem-mb  = 1176.7,
  sca2.weights-mb    = 193.5,
}

\dmrecord{g2ba}{
  sca2.abs-diff         = 0.135,
  sca2.abs-diff@fmt     = \markfst,
  sca2.abs-rel          = 6.8,
  sca2.abs-rel@fmt      = \markfst,
  sca2.coverage         = 65.9,
  sca2.delta-125        = 94.5,
  sca2.delta-125@fmt    = \markfst,
  sca2.fps              = 91.1,
  sca2.fps@fmt          = \markfst,
  sca2.lpips            = 0.473,
  sca2.peak-inf-mb      = 88.1,
  sca2.peak-inf-mb@fmt  = \markfst,
  sca2.psnr             = 23.03,
  sca2.ssim             = 0.740,
  sca2.total-mem-mb     = 115.4,
  sca2.total-mem-mb@fmt = \markfst,
  sca2.weights-mb       = 27.3,
  sca2.weights-mb@fmt   = \markfst,
}

\dmrecord{gt}{
}

\dmrecord{monosplat}{
  sca2.abs-diff         = 0.329,
  sca2.abs-rel          = 18.4,
  sca2.coverage         = 94.8,
  sca2.coverage@fmt     = \marksnd,
  sca2.delta-125        = 69.4,
  sca2.fps              = 36.6,
  sca2.fps@fmt          = \marksnd,
  sca2.lpips            = 0.284,
  sca2.peak-inf-mb      = 610.9,
  sca2.psnr             = 22.38,
  sca2.ssim             = 0.791,
  sca2.total-mem-mb     = 745.4,
  sca2.total-mem-mb@fmt = \marksnd,
  sca2.weights-mb       = 134.5,
}

\dmrecord{mvsplat}{
  sca2.abs-diff       = 0.362,
  sca2.abs-rel        = 18.8,
  sca2.coverage       = 94.6,
  sca2.delta-125      = 61.2,
  sca2.fps            = 9.9,
  sca2.lpips          = 0.243,
  sca2.lpips@fmt      = \marksnd,
  sca2.peak-inf-mb    = 3120.0,
  sca2.psnr           = 25.70,
  sca2.psnr@fmt       = \marksnd,
  sca2.ssim           = 0.810,
  sca2.ssim@fmt       = \marksnd,
  sca2.total-mem-mb   = 3165.6,
  sca2.weights-mb     = 45.6,
  sca2.weights-mb@fmt = \marksnd,
}

\dmrecord{surfelsplat}{
  sca2.abs-diff     = 0.329,
  sca2.abs-rel      = 19.0,
  sca2.coverage     = 94.6,
  sca2.delta-125    = 70.6,
  sca2.fps          = 5.0,
  sca2.lpips        = 0.485,
  sca2.peak-inf-mb  = 10351.4,
  sca2.psnr         = 16.61,
  sca2.ssim         = 0.599,
  sca2.total-mem-mb = 12385.5,
  sca2.weights-mb   = 2034.1,
}

\dmrecord{c3g}{
  sca3.abs-diff        = 0.287,
  sca3.abs-rel         = 14.6,
  sca3.coverage        = 92.1,
  sca3.delta-125       = 80.4,
  sca3.fps             = 14.6,
  sca3.lpips           = 0.510,
  sca3.peak-inf-mb     = 332.1,
  sca3.peak-inf-mb@fmt = \marksnd,
  sca3.psnr            = 18.31,
  sca3.ssim            = 0.691,
  sca3.total-mem-mb    = 4197.0,
  sca3.weights-mb      = 3864.9,
}

\dmrecord{freesplat}{
  sca3.abs-diff      = 0.187,
  sca3.abs-diff@fmt  = \marksnd,
  sca3.abs-rel       = 8.8,
  sca3.abs-rel@fmt   = \marksnd,
  sca3.coverage      = 94.5,
  sca3.coverage@fmt  = \markfst,
  sca3.delta-125     = 90.8,
  sca3.delta-125@fmt = \marksnd,
  sca3.fps           = 10.4,
  sca3.lpips         = 0.241,
  sca3.lpips@fmt     = \markfst,
  sca3.peak-inf-mb   = 1480.2,
  sca3.psnr          = 26.57,
  sca3.psnr@fmt      = \markfst,
  sca3.ssim          = 0.821,
  sca3.ssim@fmt      = \markfst,
  sca3.total-mem-mb  = 1673.7,
  sca3.weights-mb    = 193.5,
}

\dmrecord{g2ba}{
  sca3.abs-diff         = 0.160,
  sca3.abs-diff@fmt     = \markfst,
  sca3.abs-rel          = 7.7,
  sca3.abs-rel@fmt      = \markfst,
  sca3.coverage         = 73.1,
  sca3.delta-125        = 93.5,
  sca3.delta-125@fmt    = \markfst,
  sca3.fps              = 72.2,
  sca3.fps@fmt          = \markfst,
  sca3.lpips            = 0.472,
  sca3.peak-inf-mb      = 175.9,
  sca3.peak-inf-mb@fmt  = \markfst,
  sca3.psnr             = 22.71,
  sca3.ssim             = 0.750,
  sca3.total-mem-mb     = 203.2,
  sca3.total-mem-mb@fmt = \markfst,
  sca3.weights-mb       = 27.3,
  sca3.weights-mb@fmt   = \markfst,
}

\dmrecord{gt}{
}

\dmrecord{monosplat}{
  sca3.abs-diff         = 0.290,
  sca3.abs-rel          = 14.8,
  sca3.coverage         = 94.4,
  sca3.coverage@fmt     = \marksnd,
  sca3.delta-125        = 78.4,
  sca3.fps              = 30.4,
  sca3.fps@fmt          = \marksnd,
  sca3.lpips            = 0.295,
  sca3.lpips@fmt        = \marksnd,
  sca3.peak-inf-mb      = 916.1,
  sca3.psnr             = 24.08,
  sca3.psnr@fmt         = \marksnd,
  sca3.ssim             = 0.801,
  sca3.ssim@fmt         = \marksnd,
  sca3.total-mem-mb     = 1050.6,
  sca3.total-mem-mb@fmt = \marksnd,
  sca3.weights-mb       = 134.5,
}

\dmrecord{mvsplat}{
  sca3.abs-diff       = 0.432,
  sca3.abs-rel        = 21.1,
  sca3.coverage       = 94.2,
  sca3.delta-125      = 73.6,
  sca3.fps            = 5.3,
  sca3.lpips          = 0.309,
  sca3.peak-inf-mb    = 4878.0,
  sca3.psnr           = 23.77,
  sca3.ssim           = 0.787,
  sca3.total-mem-mb   = 4923.6,
  sca3.weights-mb     = 45.6,
  sca3.weights-mb@fmt = \marksnd,
}

\dmrecord{surfelsplat}{
  sca3.abs-diff     = 0.340,
  sca3.abs-rel      = 17.5,
  sca3.coverage     = 94.1,
  sca3.delta-125    = 74.6,
  sca3.fps          = 2.4,
  sca3.lpips        = 0.514,
  sca3.peak-inf-mb  = 15527.1,
  sca3.psnr         = 15.88,
  sca3.ssim         = 0.585,
  sca3.total-mem-mb = 17561.2,
  sca3.weights-mb   = 2034.1,
}

\dmrecord{c3g}{
  poisson-0-2-9.accuracy        = 11.32,
  poisson-0-2-9.chamfer         = 13.20,
  poisson-0-2-9.chamfer.scan105 = 20.00,
  poisson-0-2-9.chamfer.scan106 = 11.97,
  poisson-0-2-9.chamfer.scan110 = 8.60,
  poisson-0-2-9.chamfer.scan114 = 9.76,
  poisson-0-2-9.chamfer.scan118 = 20.00,
  poisson-0-2-9.chamfer.scan122 = 20.00,
  poisson-0-2-9.chamfer.scan24  = 12.04,
  poisson-0-2-9.chamfer.scan37  = 20.00,
  poisson-0-2-9.chamfer.scan40  = 8.23,
  poisson-0-2-9.chamfer.scan55  = 9.24,
  poisson-0-2-9.chamfer.scan63  = 12.51,
  poisson-0-2-9.chamfer.scan65  = 7.74,
  poisson-0-2-9.chamfer.scan69  = 20.00,
  poisson-0-2-9.chamfer.scan83  = 7.72,
  poisson-0-2-9.chamfer.scan97  = 10.17,
  poisson-0-2-9.completeness    = 15.08,
  poisson-0-2-9.peak-inf-mb     = 332.1,
  poisson-0-2-9.peak-inf-mb@fmt = \marksnd,
  poisson-0-2-9.scan-fps        = 14.3,
  poisson-0-2-9.total-mem-mb    = 4197.0,
  poisson-0-2-9.weights-mb      = 3864.9,
}

\dmrecord{freesplat}{
  poisson-0-2-9.accuracy        = 5.48,
  poisson-0-2-9.chamfer         = 5.06,
  poisson-0-2-9.chamfer.scan105 = 4.75,
  poisson-0-2-9.chamfer.scan106 = 4.65,
  poisson-0-2-9.chamfer.scan110 = 6.38,
  poisson-0-2-9.chamfer.scan114 = 3.32,
  poisson-0-2-9.chamfer.scan118 = 4.91,
  poisson-0-2-9.chamfer.scan122 = 6.07,
  poisson-0-2-9.chamfer.scan24  = 3.85,
  poisson-0-2-9.chamfer.scan37  = 6.25,
  poisson-0-2-9.chamfer.scan40  = 5.63,
  poisson-0-2-9.chamfer.scan55  = 4.73,
  poisson-0-2-9.chamfer.scan63  = 5.33,
  poisson-0-2-9.chamfer.scan65  = 5.49,
  poisson-0-2-9.chamfer.scan69  = 5.15,
  poisson-0-2-9.chamfer.scan83  = 5.64,
  poisson-0-2-9.chamfer.scan97  = 3.70,
  poisson-0-2-9.completeness    = 4.63,
  poisson-0-2-9.peak-inf-mb     = 1480.2,
  poisson-0-2-9.scan-fps        = 10.0,
  poisson-0-2-9.total-mem-mb    = 1673.7,
  poisson-0-2-9.weights-mb      = 193.5,
}

\dmrecord{g2ba}{
  poisson-0-2-9.accuracy            = 3.11,
  poisson-0-2-9.accuracy@fmt        = \marksnd,
  poisson-0-2-9.chamfer             = 3.50,
  poisson-0-2-9.chamfer.scan105     = 3.22,
  poisson-0-2-9.chamfer.scan106     = 3.63,
  poisson-0-2-9.chamfer.scan106@fmt = \marksnd,
  poisson-0-2-9.chamfer.scan110     = 3.63,
  poisson-0-2-9.chamfer.scan114     = 1.90,
  poisson-0-2-9.chamfer.scan114@fmt = \marksnd,
  poisson-0-2-9.chamfer.scan118     = 3.93,
  poisson-0-2-9.chamfer.scan122     = 3.42,
  poisson-0-2-9.chamfer.scan122@fmt = \marksnd,
  poisson-0-2-9.chamfer.scan24      = 4.63,
  poisson-0-2-9.chamfer.scan37      = 4.53,
  poisson-0-2-9.chamfer.scan37@fmt  = \marksnd,
  poisson-0-2-9.chamfer.scan40      = 3.41,
  poisson-0-2-9.chamfer.scan40@fmt  = \marksnd,
  poisson-0-2-9.chamfer.scan55      = 2.71,
  poisson-0-2-9.chamfer.scan55@fmt  = \marksnd,
  poisson-0-2-9.chamfer.scan63      = 3.98,
  poisson-0-2-9.chamfer.scan63@fmt  = \marksnd,
  poisson-0-2-9.chamfer.scan65      = 3.92,
  poisson-0-2-9.chamfer.scan65@fmt  = \marksnd,
  poisson-0-2-9.chamfer.scan69      = 3.02,
  poisson-0-2-9.chamfer.scan69@fmt  = \marksnd,
  poisson-0-2-9.chamfer.scan83      = 3.65,
  poisson-0-2-9.chamfer.scan83@fmt  = \marksnd,
  poisson-0-2-9.chamfer.scan97      = 2.91,
  poisson-0-2-9.chamfer.scan97@fmt  = \marksnd,
  poisson-0-2-9.chamfer@fmt         = \marksnd,
  poisson-0-2-9.completeness        = 3.89,
  poisson-0-2-9.peak-inf-mb         = 176.4,
  poisson-0-2-9.peak-inf-mb@fmt     = \markfst,
  poisson-0-2-9.scan-fps            = 69.6,
  poisson-0-2-9.scan-fps@fmt        = \markfst,
  poisson-0-2-9.total-mem-mb        = 203.7,
  poisson-0-2-9.total-mem-mb@fmt    = \markfst,
  poisson-0-2-9.weights-mb          = 27.3,
  poisson-0-2-9.weights-mb@fmt      = \markfst,
}

\dmrecord{gt}{
}

\dmrecord{monosplat}{
  poisson-0-2-9.accuracy            = 4.21,
  poisson-0-2-9.chamfer             = 3.62,
  poisson-0-2-9.chamfer.scan105     = 2.96,
  poisson-0-2-9.chamfer.scan105@fmt = \marksnd,
  poisson-0-2-9.chamfer.scan106     = 4.05,
  poisson-0-2-9.chamfer.scan110     = 3.53,
  poisson-0-2-9.chamfer.scan110@fmt = \marksnd,
  poisson-0-2-9.chamfer.scan114     = 2.42,
  poisson-0-2-9.chamfer.scan118     = 3.32,
  poisson-0-2-9.chamfer.scan118@fmt = \marksnd,
  poisson-0-2-9.chamfer.scan122     = 3.56,
  poisson-0-2-9.chamfer.scan24      = 3.59,
  poisson-0-2-9.chamfer.scan24@fmt  = \marksnd,
  poisson-0-2-9.chamfer.scan37      = 5.00,
  poisson-0-2-9.chamfer.scan40      = 3.63,
  poisson-0-2-9.chamfer.scan55      = 2.94,
  poisson-0-2-9.chamfer.scan63      = 4.90,
  poisson-0-2-9.chamfer.scan65      = 4.06,
  poisson-0-2-9.chamfer.scan69      = 3.17,
  poisson-0-2-9.chamfer.scan83      = 3.90,
  poisson-0-2-9.chamfer.scan97      = 3.33,
  poisson-0-2-9.completeness        = 3.04,
  poisson-0-2-9.completeness@fmt    = \marksnd,
  poisson-0-2-9.peak-inf-mb         = 916.1,
  poisson-0-2-9.scan-fps            = 29.1,
  poisson-0-2-9.scan-fps@fmt        = \marksnd,
  poisson-0-2-9.total-mem-mb        = 1050.6,
  poisson-0-2-9.total-mem-mb@fmt    = \marksnd,
  poisson-0-2-9.weights-mb          = 134.5,
}

\dmrecord{mvsplat}{
  poisson-0-2-9.accuracy        = 8.13,
  poisson-0-2-9.chamfer         = 6.88,
  poisson-0-2-9.chamfer.scan105 = 7.51,
  poisson-0-2-9.chamfer.scan106 = 6.60,
  poisson-0-2-9.chamfer.scan110 = 7.45,
  poisson-0-2-9.chamfer.scan114 = 8.26,
  poisson-0-2-9.chamfer.scan118 = 6.53,
  poisson-0-2-9.chamfer.scan122 = 6.21,
  poisson-0-2-9.chamfer.scan24  = 6.46,
  poisson-0-2-9.chamfer.scan37  = 7.26,
  poisson-0-2-9.chamfer.scan40  = 6.04,
  poisson-0-2-9.chamfer.scan55  = 6.41,
  poisson-0-2-9.chamfer.scan63  = 9.47,
  poisson-0-2-9.chamfer.scan65  = 4.76,
  poisson-0-2-9.chamfer.scan69  = 6.12,
  poisson-0-2-9.chamfer.scan83  = 7.20,
  poisson-0-2-9.chamfer.scan97  = 6.96,
  poisson-0-2-9.completeness    = 5.64,
  poisson-0-2-9.peak-inf-mb     = 7621.9,
  poisson-0-2-9.scan-fps        = 2.7,
  poisson-0-2-9.total-mem-mb    = 7667.5,
  poisson-0-2-9.weights-mb      = 45.6,
  poisson-0-2-9.weights-mb@fmt  = \marksnd,
}

\dmrecord{surfelsplat}{
  poisson-0-2-9.accuracy            = 1.05,
  poisson-0-2-9.accuracy@fmt        = \markfst,
  poisson-0-2-9.chamfer             = 1.62,
  poisson-0-2-9.chamfer.scan105     = 1.47,
  poisson-0-2-9.chamfer.scan105@fmt = \markfst,
  poisson-0-2-9.chamfer.scan106     = 1.91,
  poisson-0-2-9.chamfer.scan106@fmt = \markfst,
  poisson-0-2-9.chamfer.scan110     = 1.08,
  poisson-0-2-9.chamfer.scan110@fmt = \markfst,
  poisson-0-2-9.chamfer.scan114     = 1.30,
  poisson-0-2-9.chamfer.scan114@fmt = \markfst,
  poisson-0-2-9.chamfer.scan118     = 2.18,
  poisson-0-2-9.chamfer.scan118@fmt = \markfst,
  poisson-0-2-9.chamfer.scan122     = 1.88,
  poisson-0-2-9.chamfer.scan122@fmt = \markfst,
  poisson-0-2-9.chamfer.scan24      = 1.53,
  poisson-0-2-9.chamfer.scan24@fmt  = \markfst,
  poisson-0-2-9.chamfer.scan37      = 1.63,
  poisson-0-2-9.chamfer.scan37@fmt  = \markfst,
  poisson-0-2-9.chamfer.scan40      = 1.66,
  poisson-0-2-9.chamfer.scan40@fmt  = \markfst,
  poisson-0-2-9.chamfer.scan55      = 1.16,
  poisson-0-2-9.chamfer.scan55@fmt  = \markfst,
  poisson-0-2-9.chamfer.scan63      = 1.38,
  poisson-0-2-9.chamfer.scan63@fmt  = \markfst,
  poisson-0-2-9.chamfer.scan65      = 1.96,
  poisson-0-2-9.chamfer.scan65@fmt  = \markfst,
  poisson-0-2-9.chamfer.scan69      = 1.82,
  poisson-0-2-9.chamfer.scan69@fmt  = \markfst,
  poisson-0-2-9.chamfer.scan83      = 1.87,
  poisson-0-2-9.chamfer.scan83@fmt  = \markfst,
  poisson-0-2-9.chamfer.scan97      = 1.52,
  poisson-0-2-9.chamfer.scan97@fmt  = \markfst,
  poisson-0-2-9.chamfer@fmt         = \markfst,
  poisson-0-2-9.completeness        = 2.19,
  poisson-0-2-9.completeness@fmt    = \markfst,
  poisson-0-2-9.peak-inf-mb         = 15527.1,
  poisson-0-2-9.scan-fps            = 2.3,
  poisson-0-2-9.total-mem-mb        = 17561.2,
  poisson-0-2-9.weights-mb          = 2034.1,
}

\dmrecord{c3g}{
  tsdf-23-24-33.weights-mb = 3864.9,
}

\dmrecord{freesplat}{
  tsdf-23-24-33.accuracy            = 3.27,
  tsdf-23-24-33.accuracy@fmt        = \marksnd,
  tsdf-23-24-33.chamfer             = 3.78,
  tsdf-23-24-33.chamfer.scan105     = 2.86,
  tsdf-23-24-33.chamfer.scan105@fmt = \markfst,
  tsdf-23-24-33.chamfer.scan106     = 2.88,
  tsdf-23-24-33.chamfer.scan106@fmt = \markfst,
  tsdf-23-24-33.chamfer.scan110     = 3.80,
  tsdf-23-24-33.chamfer.scan110@fmt = \marksnd,
  tsdf-23-24-33.chamfer.scan114     = 2.86,
  tsdf-23-24-33.chamfer.scan114@fmt = \marksnd,
  tsdf-23-24-33.chamfer.scan118     = 3.86,
  tsdf-23-24-33.chamfer.scan118@fmt = \marksnd,
  tsdf-23-24-33.chamfer.scan122     = 3.68,
  tsdf-23-24-33.chamfer.scan24      = 3.56,
  tsdf-23-24-33.chamfer.scan24@fmt  = \markfst,
  tsdf-23-24-33.chamfer.scan37      = 5.86,
  tsdf-23-24-33.chamfer.scan37@fmt  = \marksnd,
  tsdf-23-24-33.chamfer.scan40      = 4.21,
  tsdf-23-24-33.chamfer.scan40@fmt  = \marksnd,
  tsdf-23-24-33.chamfer.scan55      = 3.22,
  tsdf-23-24-33.chamfer.scan55@fmt  = \marksnd,
  tsdf-23-24-33.chamfer.scan63      = 5.43,
  tsdf-23-24-33.chamfer.scan65      = 4.40,
  tsdf-23-24-33.chamfer.scan69      = 3.13,
  tsdf-23-24-33.chamfer.scan69@fmt  = \marksnd,
  tsdf-23-24-33.chamfer.scan83      = 3.52,
  tsdf-23-24-33.chamfer.scan83@fmt  = \markfst,
  tsdf-23-24-33.chamfer.scan97      = 3.39,
  tsdf-23-24-33.chamfer.scan97@fmt  = \marksnd,
  tsdf-23-24-33.chamfer@fmt         = \marksnd,
  tsdf-23-24-33.completeness        = 4.28,
  tsdf-23-24-33.completeness@fmt    = \marksnd,
  tsdf-23-24-33.peak-inf-mb         = 1480.2,
  tsdf-23-24-33.scan-fps            = 9.7,
  tsdf-23-24-33.total-mem-mb        = 1673.7,
  tsdf-23-24-33.weights-mb          = 193.5,
}

\dmrecord{g2ba}{
  tsdf-23-24-33.accuracy            = 2.73,
  tsdf-23-24-33.accuracy@fmt        = \markfst,
  tsdf-23-24-33.chamfer             = 3.23,
  tsdf-23-24-33.chamfer.scan105     = 3.16,
  tsdf-23-24-33.chamfer.scan105@fmt = \marksnd,
  tsdf-23-24-33.chamfer.scan106     = 3.88,
  tsdf-23-24-33.chamfer.scan106@fmt = \marksnd,
  tsdf-23-24-33.chamfer.scan110     = 3.70,
  tsdf-23-24-33.chamfer.scan110@fmt = \markfst,
  tsdf-23-24-33.chamfer.scan114     = 1.98,
  tsdf-23-24-33.chamfer.scan114@fmt = \markfst,
  tsdf-23-24-33.chamfer.scan118     = 3.35,
  tsdf-23-24-33.chamfer.scan118@fmt = \markfst,
  tsdf-23-24-33.chamfer.scan122     = 2.67,
  tsdf-23-24-33.chamfer.scan122@fmt = \markfst,
  tsdf-23-24-33.chamfer.scan24      = 4.39,
  tsdf-23-24-33.chamfer.scan37      = 4.03,
  tsdf-23-24-33.chamfer.scan37@fmt  = \markfst,
  tsdf-23-24-33.chamfer.scan40      = 2.69,
  tsdf-23-24-33.chamfer.scan40@fmt  = \markfst,
  tsdf-23-24-33.chamfer.scan55      = 2.19,
  tsdf-23-24-33.chamfer.scan55@fmt  = \markfst,
  tsdf-23-24-33.chamfer.scan63      = 3.35,
  tsdf-23-24-33.chamfer.scan63@fmt  = \markfst,
  tsdf-23-24-33.chamfer.scan65      = 3.03,
  tsdf-23-24-33.chamfer.scan65@fmt  = \markfst,
  tsdf-23-24-33.chamfer.scan69      = 2.88,
  tsdf-23-24-33.chamfer.scan69@fmt  = \markfst,
  tsdf-23-24-33.chamfer.scan83      = 3.84,
  tsdf-23-24-33.chamfer.scan83@fmt  = \marksnd,
  tsdf-23-24-33.chamfer.scan97      = 3.28,
  tsdf-23-24-33.chamfer.scan97@fmt  = \markfst,
  tsdf-23-24-33.chamfer@fmt         = \markfst,
  tsdf-23-24-33.completeness        = 3.73,
  tsdf-23-24-33.completeness@fmt    = \markfst,
  tsdf-23-24-33.peak-inf-mb         = 176.4,
  tsdf-23-24-33.peak-inf-mb@fmt     = \markfst,
  tsdf-23-24-33.scan-fps            = 68.9,
  tsdf-23-24-33.scan-fps@fmt        = \markfst,
  tsdf-23-24-33.total-mem-mb        = 203.7,
  tsdf-23-24-33.total-mem-mb@fmt    = \markfst,
  tsdf-23-24-33.weights-mb          = 27.3,
  tsdf-23-24-33.weights-mb@fmt      = \markfst,
}

\dmrecord{gt}{
}

\dmrecord{monosplat}{
  tsdf-23-24-33.accuracy            = 3.43,
  tsdf-23-24-33.chamfer             = 4.19,
  tsdf-23-24-33.chamfer.scan105     = 3.58,
  tsdf-23-24-33.chamfer.scan106     = 4.00,
  tsdf-23-24-33.chamfer.scan110     = 4.08,
  tsdf-23-24-33.chamfer.scan114     = 3.24,
  tsdf-23-24-33.chamfer.scan118     = 4.24,
  tsdf-23-24-33.chamfer.scan122     = 3.68,
  tsdf-23-24-33.chamfer.scan122@fmt = \marksnd,
  tsdf-23-24-33.chamfer.scan24      = 3.91,
  tsdf-23-24-33.chamfer.scan24@fmt  = \marksnd,
  tsdf-23-24-33.chamfer.scan37      = 6.56,
  tsdf-23-24-33.chamfer.scan40      = 5.09,
  tsdf-23-24-33.chamfer.scan55      = 3.82,
  tsdf-23-24-33.chamfer.scan63      = 5.41,
  tsdf-23-24-33.chamfer.scan63@fmt  = \marksnd,
  tsdf-23-24-33.chamfer.scan65      = 3.87,
  tsdf-23-24-33.chamfer.scan65@fmt  = \marksnd,
  tsdf-23-24-33.chamfer.scan69      = 3.26,
  tsdf-23-24-33.chamfer.scan83      = 4.16,
  tsdf-23-24-33.chamfer.scan97      = 3.98,
  tsdf-23-24-33.completeness        = 4.96,
  tsdf-23-24-33.peak-inf-mb         = 916.1,
  tsdf-23-24-33.peak-inf-mb@fmt     = \marksnd,
  tsdf-23-24-33.scan-fps            = 29.9,
  tsdf-23-24-33.scan-fps@fmt        = \marksnd,
  tsdf-23-24-33.total-mem-mb        = 1050.6,
  tsdf-23-24-33.total-mem-mb@fmt    = \marksnd,
  tsdf-23-24-33.weights-mb          = 134.5,
}

\dmrecord{mvsplat}{
  tsdf-23-24-33.accuracy        = 10.17,
  tsdf-23-24-33.chamfer         = 10.56,
  tsdf-23-24-33.chamfer.scan105 = 10.54,
  tsdf-23-24-33.chamfer.scan106 = 8.84,
  tsdf-23-24-33.chamfer.scan110 = 12.33,
  tsdf-23-24-33.chamfer.scan114 = 9.89,
  tsdf-23-24-33.chamfer.scan118 = 8.45,
  tsdf-23-24-33.chamfer.scan122 = 9.56,
  tsdf-23-24-33.chamfer.scan24  = 12.96,
  tsdf-23-24-33.chamfer.scan37  = 10.65,
  tsdf-23-24-33.chamfer.scan40  = 8.16,
  tsdf-23-24-33.chamfer.scan55  = 8.62,
  tsdf-23-24-33.chamfer.scan63  = 13.31,
  tsdf-23-24-33.chamfer.scan65  = 9.59,
  tsdf-23-24-33.chamfer.scan69  = 12.25,
  tsdf-23-24-33.chamfer.scan83  = 11.55,
  tsdf-23-24-33.chamfer.scan97  = 11.65,
  tsdf-23-24-33.completeness    = 10.94,
  tsdf-23-24-33.peak-inf-mb     = 7621.9,
  tsdf-23-24-33.scan-fps        = 2.7,
  tsdf-23-24-33.total-mem-mb    = 7667.5,
  tsdf-23-24-33.weights-mb      = 45.6,
  tsdf-23-24-33.weights-mb@fmt  = \marksnd,
}

\dmrecord{surfelsplat}{
  tsdf-23-24-33.accuracy        = 7.56,
  tsdf-23-24-33.chamfer         = 8.15,
  tsdf-23-24-33.chamfer.scan105 = 6.26,
  tsdf-23-24-33.chamfer.scan106 = 7.68,
  tsdf-23-24-33.chamfer.scan110 = 6.70,
  tsdf-23-24-33.chamfer.scan114 = 7.09,
  tsdf-23-24-33.chamfer.scan118 = 7.10,
  tsdf-23-24-33.chamfer.scan122 = 5.77,
  tsdf-23-24-33.chamfer.scan24  = 9.00,
  tsdf-23-24-33.chamfer.scan37  = 8.42,
  tsdf-23-24-33.chamfer.scan40  = 8.05,
  tsdf-23-24-33.chamfer.scan55  = 5.94,
  tsdf-23-24-33.chamfer.scan63  = 12.88,
  tsdf-23-24-33.chamfer.scan65  = 6.29,
  tsdf-23-24-33.chamfer.scan69  = 8.20,
  tsdf-23-24-33.chamfer.scan83  = 11.60,
  tsdf-23-24-33.chamfer.scan97  = 11.22,
  tsdf-23-24-33.completeness    = 8.73,
  tsdf-23-24-33.peak-inf-mb     = 15527.1,
  tsdf-23-24-33.scan-fps        = 2.4,
  tsdf-23-24-33.total-mem-mb    = 17561.2,
  tsdf-23-24-33.weights-mb      = 2034.1,
}

\newcommand{\tpmin}{\dmeval[0]{\dmv{g2ba}{tsdf-23-24-33.scan-fps}}}
\newcommand{\tpmax}{\dmeval[0]{\dmv{g2ba}{rep2.fps}}}

\newcommand{\memmin}{\dmeval[0]{\dmv{g2ba}{rep2.total-mem-mb}}}
\newcommand{\memmax}{\dmeval[0]{\dmv{g2ba}{rep3.total-mem-mb}}}

\newcommand{\memredxmin}{\dmeval[0]{\dmv{monosplat}{tsdf-23-24-33.total-mem-mb} / \dmv{g2ba}{tsdf-23-24-33.total-mem-mb}}}
\newcommand{\memredxmax}{\dmeval[0]{\dmv{surfelsplat}{rep2.total-mem-mb} / \dmv{g2ba}{rep2.total-mem-mb}}}

\newcommand{\speedupxmin}{\dmeval[1]{\dmv{g2ba}{tsdf-23-24-33.scan-fps} / \dmv{monosplat}{tsdf-23-24-33.scan-fps}}}
\newcommand{\speedupxmax}{\dmeval[0]{\dmv{g2ba}{sca2.fps} / \dmv{surfelsplat}{rep2.fps}}}

\pgfplotsset{compat=1.18}

\pgfplotsset{colormap={turbo}{
    rgb255(0cm)=(48,18,59)
    rgb255(1cm)=(69,78,162)
    rgb255(2cm)=(38,130,217)
    rgb255(3cm)=(31,177,185)
    rgb255(4cm)=(78,206,128)
    rgb255(5cm)=(167,222,60)
    rgb255(6cm)=(229,218,43)
    rgb255(7cm)=(247,163,26)
    rgb255(8cm)=(230,90,7)
    rgb255(9cm)=(184,39,4)
    rgb255(10cm)=(122,4,3)
}}

\pgfplotsset{colormap={inferno}{
    rgb(0)=(0.001462, 0.000466, 0.013866),
    rgb(15)=(0.037668, 0.025921, 0.132232),
    rgb(30)=(0.116656, 0.047574, 0.272321),
    rgb(45)=(0.217949, 0.036615, 0.383522),
    rgb(60)=(0.316282, 0.053490, 0.425116),
    rgb(75)=(0.410113, 0.087896, 0.433098),
    rgb(90)=(0.503493, 0.121575, 0.423356),
    rgb(105)=(0.596940, 0.154848, 0.398125),
    rgb(120)=(0.688653, 0.192239, 0.357603),
    rgb(135)=(0.775059, 0.239667, 0.303526),
    rgb(150)=(0.851384, 0.302260, 0.239636),
    rgb(165)=(0.912966, 0.381636, 0.169755),
    rgb(180)=(0.956852, 0.475356, 0.094695),
    rgb(195)=(0.981895, 0.579392, 0.026250),
    rgb(210)=(0.987464, 0.690366, 0.079990),
    rgb(225)=(0.973088, 0.805409, 0.216877),
    rgb(240)=(0.947594, 0.917399, 0.410665),
    rgb(255)=(0.988362, 0.998364, 0.644924),
}}

\title{
G$^2$SR: Geometric Methods for Fast and Memory- Efficient Gaussian-based Surface Reconstruction
}

\author{
    Dasong Gao$^{\textsuperscript{\orcidicon{0000-0002-1391-0869}}}$,
    Vivienne Sze$^{\textsuperscript{\orcidicon{0000-0003-4841-3990}}}$\textit{, Senior Member, IEEE}, and
    Sertac Karaman$^{\textsuperscript{\orcidicon{0000-0002-2225-7275}}}$\textit{, Member, IEEE}

    \thanks{The authors are with the Massachusetts Institute of Technology, Cambridge, MA  02139,  USA (e-mail: \href{mailto:dasongg@mit.edu}{dasongg@mit.edu}; \href{mailto:sze@mit.edu}{sze@mit.edu}; sertac@mit.edu).}

}

\begin{document}

\maketitle

\thispagestyle{empty}
\pagestyle{empty}


\begin{abstract}
Few-view surface reconstruction recovers the visible surfaces of a scene from a few posed RGB images, providing the 3D models that robots need to explore and interact online.
On mobile platforms, the reconstruction must be fast and geometrically accurate while keeping a small memory footprint to ensure safe and efficient operation.
3D Gaussian Splatting (3DGS) offers a high-fidelity scene representation, but building it from a few views is ill-posed, as many distinct surfaces reproduce the same images, making traditional photometric methods prone to ``floater'' artifacts.
End-to-end methods resolve the ambiguity by regressing splats with large, usually Transformer-based, networks that require heavy compute and memory while generalizing poorly to new scenes.
We propose G$^2$SR, which exploits a well-posed core of the task: given cross-view 2D splat correspondences, 3D splats follow analytically from multi-view geometry.
G$^2$SR employs a lightweight neural frontend to detect and track 2D Gaussian splats on the image plane and an analytic backend to triangulate each into a metric-scale 3D splat.
On ScanNet, Replica, and DTU, G$^2$SR matches or exceeds the geometric accuracy of state-of-the-art end-to-end methods while running at \tpmin--\tpmax~reconstructions per second within \memmax~MB of GPU memory (\memredxmin--\memredxmax$\times$ less) for 2- and 3-view inputs at $384 \times 512$ resolution, offering a practical path to online Gaussian-based surface reconstruction.

\end{abstract}


\begin{figure}
    \centering
    \includegraphics[width=0.95\linewidth]{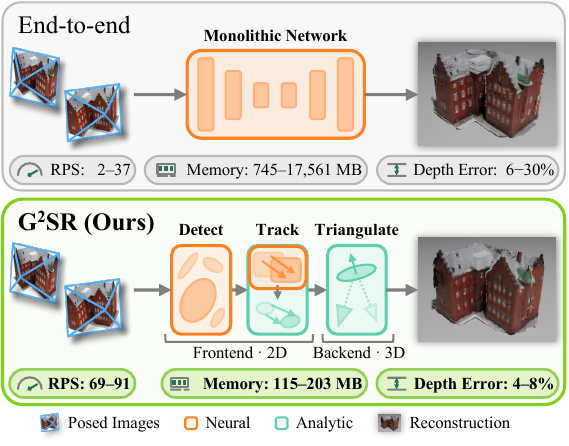} \\
    \vspace{4pt}
    \begin{tikzpicture}
        \node (fig) [anchor=south west,inner sep=0] at (0,0) {\includegraphics[width=0.94\linewidth]{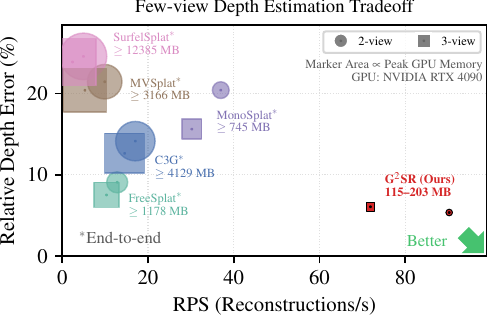}};


    \end{tikzpicture}
    \vspace{-6pt}
    \caption{
    From a few (2--3) posed RGB images, prior end-to-end methods (top) solve the ill-posed reconstruction task as a whole by regressing 3D Gaussian splats with a single monolithic network that requires significant compute and memory but generalizes poorly. G$^2$SR (middle) instead decouples the task: a small neural \emph{frontend} learns the ill-posed 2D subproblems of \textbf{detecting} 2D Gaussian splats and \textbf{tracking} their cross-view correspondences, while an analytic \emph{backend} solves the well-posed 3D subproblem of \textbf{triangulating} each splat from multi-view geometry. The scatter plot (bottom) summarizes 2- and 3-view depth estimation averaged over Replica~\cite{straub2019replica} and ScanNet~\cite{dai2017scannet}: G$^2$SR achieves lower depth error at higher throughput (RPS) while using a fraction of the baselines' GPU memory (marker area).
    }
    \label{fig:tradeoff}
    \vspace{-18pt}
\end{figure}

\section{Introduction}

\IEEEPARstart{F}{\MakeUppercase{ew-view}} surface reconstruction, which recovers the scene's visible surfaces from a few posed RGB images, provides the essential 3D modeling required for various applications including autonomous exploration, augmented reality, and industrial inspection.
In these applications, robots need to build the scene \textit{online} as they travel and quickly decide how to act based on its structure.
The throughput and geometric accuracy of the reconstruction algorithm are therefore critical for robots' fast and safe interaction with the environment.

As these tasks usually run on platforms with limited on-board compute and battery, such as small drones and wearable devices, surface reconstruction algorithms must also maintain a small memory footprint (the maximum memory in use during execution) to be effective.
Beyond competing for a fixed capacity with other concurrent tasks such as localization, a high-memory reconstruction algorithm must frequently transmit data to and from the off-chip DRAM, which could dominate the system compute energy~\cite{gmmap} and introduce delays to prevent full utilization of the compute units~\cite{liu2023efficientvit}.
Therefore, surface reconstruction algorithms must also limit their memory footprint to support long-term and efficient operation.

Recently, 3D Gaussian Splatting (3DGS) \cite{kerbl20233d, huang20242d, guedon2024sugar, chen2024pgsr} has emerged as a compelling candidate scene representation due to its high fidelity and efficient view recovery through rendering.
However, quickly and accurately reconstructing such a representation from a few views with a low memory footprint remains challenging because few-view reconstruction is fundamentally \textit{ill-posed} and requires additional compute and memory to resolve the ambiguity.
From only a few views, many distinct 3D surfaces reproduce almost identical images, which makes it difficult to infer the true ones from the captured images alone.
As a result, traditional 3DGS \cite{kerbl20233d, huang20242d, guedon2024sugar, chen2024pgsr}, which fits Gaussian splats by matching their rendering to the captured images, struggles with few views by producing ``floater'' artifacts and erroneous depth~\cite{wu2025sparse2dgs, huang2025fatesgs, gao2025gevo}.

To resolve ambiguity, an emerging family of end-to-end methods~\cite{chen2024mvsplat, wang2024freesplat, liu2025monosplat, an2025c3g, daisurfelsplat} leverages neural networks to directly regress the splat parameters from input images.
Trained on various scenes, the networks semantically understand the scene structure and assign plausible positions to each splat with rules learned from data.
However, encoding large volumes of prior knowledge on 3D scene structures typically requires heavy backbones such as Vision Transformers (ViTs), which need hundreds of MBs to tens of GBs of memory for weights and activations and take up to seconds to complete a single inference on high-end desktop GPUs.
Moreover, the networks are usually overfit to the scale, texture, and camera configurations of the training set and predict erroneous depth when test views differ in those aspects.
This prohibitive resource usage and poor generalization, resulting from attacking the whole \textit{ill-posed} problem with a monolithic network, has limited their potential for on-board reconstruction.

In this letter, we resolve these issues by exploiting a \textit{well-posed} subproblem within the otherwise \textit{ill-posed} task: given cross-view 2D splat correspondences, 3D splats analytically follow from classical multi-view geometry.
Unlike a deep network approximating the same transformation, triangulation solves 3D splats from the projective constraints of their own correspondences rather than recalling prior knowledge learned from other scenes;
costing a few linear algebra operations per splat, it recovers the metric scale and automatically adapts to varying cameras.
Identifying corresponding splats remains \textit{ill-posed} and benefits from a neural prior, but the networks can stay lightweight while remaining generalizable: they only compare images to track 2D texture and need no capacity to memorize 3D structure or approximate multi-view geometry.

Therefore, we propose G$^2$SR, an efficient Gaussian-based few-view surface reconstruction framework with a small neural frontend and an analytic backend for solving subproblems of matching posedness (\Cref{fig:tradeoff}, middle):
\begin{itemize}
    \item (\textit{Frontend, ill-posed}) a \textbf{splat detection network} partitions an input image into boundary-respecting 2D Gaussian splats;
     an \textbf{affine tracking algorithm} then estimates their correspondences across views from neural optical flow;
    \item (\textit{Backend, well-posed}) a \textbf{geometric triangulator} recovers 3D splats by matching projections to correspondences under the Hellinger distance using Gauss--Newton iterations.
\end{itemize}
On the ScanNet~\cite{dai2017scannet}, Replica~\cite{straub2019replica}, and DTU~\cite{jensen2014dtu} datasets, G$^2$SR matches or exceeds the depth-estimation and mesh reconstruction accuracy of state-of-the-art (SOTA) end-to-end methods in 2- and 3-view reconstruction.
Consistent with this design, G$^2$SR recovers depth at metric scale without needing post-hoc alignment to metric truth (\Cref{sec:exp:geom-acc:depth-estimation}), and its accuracy holds when the input camera geometry varies across datasets (\Cref{sec:exp:geom-acc:generalization}).
When measured on a single NVIDIA RTX 4090 GPU at a resolution of $384 \times 512$, G$^2$SR operates at \tpmin--\tpmax~RPS while only consuming \memmin--\memmax~MB of memory, resulting in a speedup and memory reduction of up to \speedupxmax$\times$ and \memredxmax$\times$, respectively (\Cref{fig:tradeoff}, bottom).

\section{Related Work}

\subsection{End-to-end GS}

Traditional per-scene 3D GS \cite{kerbl20233d, chen2024pgsr} fits the Gaussian parameters to each scene via iterative differentiable rendering.
Despite recovering photorealistic images at thousands of FPS \cite{kerbl20233d}, the reconstruction takes up to 30,000 optimization steps and hours per scene, and is susceptible to the ``floater'' artifacts due to view ambiguity in few-view settings \cite{daisurfelsplat}.
Both combined make these methods unsuitable for online applications.

The recent end-to-end Gaussian Splatting has opened new possibilities for near-instantaneous few-view 3D reconstruction.
By using deep neural networks to predict splat attributes decoded from unprojected per-pixel depths \cite{chen2024mvsplat, liu2024mvsgaussian, xu2025depthsplat, liu2025monosplat, lou2025mugs} or point clouds \cite{xu2025depthsplat, zhang2025transplat, hosseinzadeh2025g3splat, an2025c3g, moreau2025off}, these methods recover the 3D scene structure by leveraging prior knowledge learned from various training scenes.

However, locating and shaping Gaussians in 3D forces the networks to both encode scene priors that resolve the ambiguity and aggregate cross-view information to approximate multi-view geometry, driving current SOTA frameworks to employ ``heavy'' backbones, primarily Vision Transformers (ViTs), that incur significant memory and compute yet are hard to generalize to untrained scenes.
%
CNN-based variants such as \textit{pixelSplat} \cite{charatan2024pixelsplat} and \textit{MonoSplat} \cite{liu2025monosplat}, despite being less resource-demanding, degrade in geometric accuracy due to smaller network capacity.
In contrast, G$^2$SR uses an analytic backend to solve the well-posed triangulation subproblem from multi-view constraints, while lightweight networks supply only 2D image-plane evidence, therefore reducing compute and memory while ensuring across-scene generalization.

\begin{figure*}[t]
	\centering
	\includegraphics[width=\textwidth]{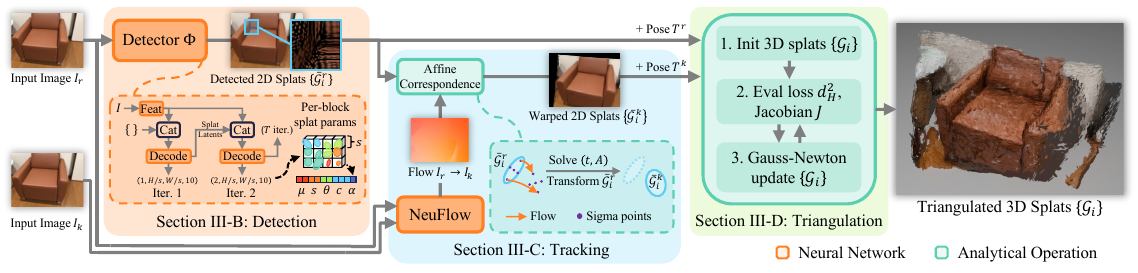}
	\caption{Overview of the G$^2$SR pipeline. A lightweight network detects 2D Gaussian splats $\{\bar{\mathcal{G}}_i^r\}$ on the reference image $I_r$ (\Cref{sec:method:detection}); an optical-flow-based sigma-point warping then establishes affine correspondences $\{\bar{\mathcal{G}}_i^k\}$ to view $k$ (\Cref{sec:method:correspondence}); finally, a Gauss--Newton solver triangulates 3D splats $\{\hat{\mathcal{G}}_i\}$ by matching projections under the squared Hellinger distance $d_H^2$ (\Cref{sec:method:triangulation}). During 3-view reconstruction, we concatenate the 3D splats generated with $r = 1, k = 2$ and $r = 2, k = 3$.}
	\label{fig:architecture}
\end{figure*}

\subsection{Multi-view Geometry in GS}

In both per-scene and end-to-end Gaussian splatting, multi-view geometry has so far served only \textit{auxiliary} roles of initialization, regularization, or post-processing.
Per-scene methods usually seed the splat centers with sparse point clouds from offline SfM \cite{kerbl20233d}.
More recent variants, such as \textit{EDGS} \cite{kotovenko2025edgs} and \textit{TriaGS} \cite{tran2025triags}, use dense triangulation to replace densification or as consensus constraints to condition the optimization under dense views.
To improve surface smoothness and mitigate the ``floater'' artifacts in few-view settings, subsequent works introduce geometric priors such as monocular depth and normal consistency to regularize the optimization \cite{chen2024pgsr, wu2025sparse2dgs, zhao2025fsfsplatter, huang2025fatesgs}.
However, even with the constraints imposed, the splats are still recovered using photometric optimization in each scene, whose long latency precludes their online use.
In the end-to-end regime, multi-view constraints are often implicitly integrated within the neural architecture via feature cross-attention in 3D cost volumes \cite{chen2024mvsplat, charatan2024pixelsplat, liu2024mvsgaussian, lou2025mugs} or among the transformer tokens, or explicitly used to prune floater Gaussians after initial inference \cite{liu2024mvsgaussian, wang2024freesplat}.
In contrast, our framework is among the first to employ analytic multi-view triangulation as the \textit{primary} 3D solver in a feed-forward pipeline, rather than as an auxiliary for seeding or regularizing per-scene optimization \cite{kotovenko2025edgs, tran2025triags} or for pruning after network inference \cite{liu2024mvsgaussian, wang2024freesplat}, delivering online, metric-scale reconstruction within hundreds of megabytes of memory.

\section{Proposed Method}\label{sec:method}

In this section, we present G$^2$SR, an efficient Gaussian-based few-view surface reconstruction framework that \textit{analytically solves} the \textit{well-posed} triangulation subproblem of the task and \textit{learns} the \textit{ill-posed} splat detection and tracking subproblems.
Our design rests on a simple observation: a Gaussian splat lying on the scene surface is fully determined by its 2D projections across the input views, thus making its position, shape, and orientation recoverable through multi-view geometry.
G$^2$SR mirrors the detect--track--optimize pipeline of feature-based SLAM (\Cref{fig:architecture}): a lightweight network \textbf{detects} 2D splats on the reference view (\Cref{sec:method:detection}) from local texture; an optical-flow-based sigma-point warping algorithm \textbf{tracks} each splat across the remaining views as an affine correspondence (\Cref{sec:method:correspondence}); and a Gauss--Newton solver \textbf{triangulates} every 3D splat from these correspondences using a few linear algebraic operations (\Cref{sec:method:triangulation}), so that geometric accuracy and generalization rest on a multi-view-constrained-backend, making lightweight networks suffice in the frontend.
Before detailing each stage, we first introduce notations and formalize the projection-matching objective the backend solves (\Cref{sec:method:formulation}).
\subsection{Formulation}\label{sec:method:formulation}

Given $K$ posed RGB images $I_{1\dots K}$ of a static scene, where $K = 2, 3$ in the few-view setting we target, we seek to recover a set of $N$ oriented 3D Gaussian splats to represent the visible surface.
Similar to 3DGS \cite{kerbl20233d}, we denote the geometry of each splat as $\mathcal{G} = (\mat{\mu}, \Sigma) = (\mat{\mu}, R S S^T R^T)$, which is parameterized by mean $\mat{\mu} \in \mathbb{R}^3$, rotation $R \in \mathrm{SO}(3)$, and scales $S = \diag(s_u, s_v, \varepsilon)$ with $s_u, s_v > 0$ and a fixed small positive $\varepsilon$.
Together with view-independent color\footnote{We ignore view-dependent spherical-harmonic color for simplicity.} $\mat{c} \in [0, 1]^{3}$ and opacity $\alpha \in [0, 1]$, the splat represents a colored, semi-transparent, thin elliptical disk centered at $\mat{\mu}$ with semi-axes $s_u, s_v$ and normal $R\mat{e}_z$; each splat thus models a local planar patch of the scene surface, \ie, a surfel.

During reconstruction, we recover the 3D splats through \emph{projection matching}: rather than rendering them and comparing with the captured pixels, we match each splat's 2D projection in every view against a per-view 2D target.
Specifically, let $\mathcal{G}_i^*$ denote the splats that, once rendered, reproduce each captured image by minimizing the 3DGS rendering objective
\begin{equation} \label{eqn:conventional-gs-opt}
    \left\{ \hat{\mathcal{G}}_i \right\} = \argmin_{\left\{ \mathcal{G}_i \right\}} \sum_{k \in [K]} \mathcal{L} \left( \mathcal{R} \left( \left\{ \pi^k (\mathcal{G}_i), \mat{c}_i, \alpha_i \right\} \right), I_k \right),
\end{equation}
where $\pi^k$ projects a splat into the $k$-th view, $\mathcal{R}$ renders the projected splats by alpha-blending, and $\mathcal{L}$ is the rendering loss function against the captured image $I_k$.
We seek to recover each splat by matching its projection to that of $\mathcal{G}_i^*$ in every view for some distance metric $d$ between 2D splats,
\begin{equation} \label{eq:gt-proj-opt}
    \left\{ \hat{\mathcal{G}}_i \right\} = \argmin_{\left\{ \mathcal{G}_i \right\}}\, \sum_{i \in [N]}\, \sum_{k \in [K]}\, d^2\!\left(\pi^k(\mathcal{G}_i),\, \pi^k(\mathcal{G}_i^*) \right).
\end{equation}

Matching these projections is, under idealized conditions, sufficient to reconstruct the scene.
As rendering depends on the splats only through the per-view inputs $\{\pi^k(\mathcal{G}_i), \mat{c}_i, \alpha_i\}$ that $\mathcal{R}$ consumes, any splats whose projections equal those of $\mathcal{G}_i^*$ render exactly to the images that $\mathcal{G}_i^*$ does, and thus also minimize \Cref{eqn:conventional-gs-opt}\footnote{Up to rendering order; \Cref{sec:method:detection} alleviates this constraint by training the detector to predict 2D splats that render correctly under any ordering.}.
That is, with proper per-view 2D targets, recovering the 3D splats reduces to the well-posed projection matching problem \Cref{eq:gt-proj-opt}, which we analytically solve in the backend (\Cref{sec:method:triangulation}).

In reality, however, the projections $\pi^k(\mathcal{G}_i^*)$ are neither observable nor unique: there are infinitely many scenes that reproduce the captured images, and each can be partitioned into 2D splats in infinitely many ways.
G$^2$SR therefore resolves this ill-posed subproblem by constructing one such set of 2D targets with its neural frontend by detecting them in the reference view (\Cref{sec:method:detection}) and tracking them to the remaining views (\Cref{sec:method:correspondence}).

\subsection{Neural Splat Detection}\label{sec:method:detection}

We task a lightweight network $\Phi$ with predicting, out of many configurations, a set of 2D Gaussian splats $\{\bar{\mathcal{G}}_i^r\}$ for a chosen reference image $I_r$ that, after triangulation, jointly represent the visible surfaces in $I_r$.
Like its 3D counterpart, each predicted 2D splat is an oriented Gaussian $\bar{\mathcal{G}}_i^r = (\bar{\mat{\mu}}_i^r, \bar{\Sigma}_i^r)$ with center $\bar{\mat{\mu}}_i^r \in \mathbb{R}^2$ and covariance $\bar{\Sigma}_i^r = R(\theta_i^r)\,\bar{S}_i^r\,\bar{S}_i^{rT} R(\theta_i^r)^T$, factored into a 2D rotation $R(\theta) \in \mathrm{SO}(2)$ and scales $\bar{S} = \diag(\bar{s}_u, \bar{s}_v)$.
During reconstruction, we take all but the last frame as the reference (\Cref{fig:architecture}).

Since Gaussian splats are naturally supervised via the rendering loss, we make $\Phi$ additionally predict photometric attributes $(\bar{\mat{c}}_i^r, \bar{\alpha}_i^r)$ per splat.
Formally, $\left\{ \bar{\mathcal{G}}_i^r, \bar{\mat{c}}_i^r, \bar{\alpha}_i^r \right\} = \Phi(I_r)$.

\subsubsection{Architecture}
For input image $I_r$, $\Phi$ first extracts a feature map of size $(H/s, W/s, 64)$ using a 3-layer convolutional network.
A decoder head then iterates $T$ times to predict $T$ splats per $s\times s$ block of $I_r$, where each iteration involves: 1) generating one new splat token per block from the feature map; 2) concatenating with previously generated tokens; and 3) performing in-block cross-attention to resolve incompatibilities among splats in the same block.
The tokens are mapped into a 10-dimensional vector encoding the splat parameters: $(\bar{\mat{\mu}}_i^r, \bar{s}_u, \bar{s}_v, \sin(\theta_i^r), \cos(\theta_i^r), \bar{\mat{c}}_i^r, \bar{\alpha}_i^r)$, finally resulting in a total of $T \cdot (H/s) \cdot (W/s)$ splats for the whole image (\Cref{fig:architecture}).

Unlike 3D attributes, such as depth or surface normal, that are predicted from information aggregated across multiple images, each splat predicted by $\Phi$ inspects local texture and color and requires no capacity to memorize 3D scene structure or approximate multi-view geometry.
This allows us to deploy a lightweight network with only 0.41~M parameters, which is $6.2\times$ and $53.7\times$ smaller than popular backbones such as MobileNetV3-Small and ViT-S, respectively, thus shrinking the memory footprint of neural inference (\Cref{sec:exp:efficiency:memory}).

\subsubsection{Training}
To ensure $\{\bar{\mathcal{G}}_i^r\}$ can be triangulated into an accurate 3D representation of the visible surfaces, we desire $\{\bar{\mathcal{G}}_i^r\}$ to: 1) cover most of the image, and 2) respect surface boundaries.
Since surface boundaries usually correlate with those of color, we supervise $\Phi$ end-to-end by penalizing differences in rasterized color with the ordinary $\ell_1$ and structural similarity index measure (SSIM) losses from 3DGS \cite{kerbl20233d}:
\begin{equation} \label{eq:detection-loss}
    \mathcal{L} = \lambda_1 \| \bar{I}_r - I_r \|_1 + \lambda_2 \left( 1 - {\mathrm{SSIM}} \left( \bar{I}_r,\; I_r \right) \right)
\end{equation}
where $\bar{I}_r = \mathcal{R} \left( \left\{ \bar{\mathcal{G}}_i, \bar{\mat{c}}_i, \bar{\alpha}_i \right\} \right)$ is the image rasterized from predicted splats, and $\lambda_1, \lambda_2$ are weights.

Under the rasterizer's fixed front-to-back order, \Cref{eq:detection-loss} does not effectively penalize boundary-violating splats at the end of the list due to their low visibility during alpha-blending.
Such splats break the local-planarity assumption underlying our triangulation, yielding inconsistent targets under the warping of \Cref{sec:method:correspondence}.
We therefore randomly permute the predicted splats before rasterization, forcing $\Phi$ to produce a set valid under any ordering and closing the caveat raised in \Cref{sec:method:formulation}.
We train $\Phi$ on the RealEstate10K dataset \cite{zhou2018stereo} due to its diverse distribution of image appearances.

\subsection{Affine Splat Tracking}\label{sec:method:correspondence}

After detecting a set of 2D splats $\{\bar{\mathcal{G}}_i^r\}$ that tile $I_r$, we locate the respective projection targets in other views, \ie, 2D splats $\{\bar{\mathcal{G}}_i^k\}_{k > r}$ that correspond to the same surfel in view $k$.

As a change of viewpoint translates a splat's center and deforms its contour due to perspective projection, we characterize such a cross-view correspondence by a local affine transform $(\mat{t}_i^k, A_i^k)$, with translation $\mat{t}_i^k \in \mathbb{R}^2$ and linear mapping $A_i^k \in \mathrm{GL}(2)$.
Together, the two components map the support region $\Omega(\bar{\mathcal{G}}_i^r)$ of the reference splat onto its counterpart $\Omega(\bar{\mathcal{G}}_i^k)$ in view $k$, where $\Omega(\bar{\mathcal{G}}) = \{ \mat{x} : \norm{ \mat{x} - \bar{\mat{\mu}} }_{\bar{\Sigma}} \le 2 \}$ is the elliptical region within the 2-sigma radius.
Once $(\mat{t}_i^k, A_i^k)$ is estimated, the corresponding 2D splat in view $k$ is then assembled using first-order approximation as
\begin{equation}
    \bar{\mathcal{G}}_i^k := (\bar{\mat{\mu}}_i^k, \bar{\Sigma}_i^k) = (\bar{\mat{\mu}}_i^r + \mat{t}_i^k,\; A_i^k \bar{\Sigma}_i^r (A_i^k)^T).
\end{equation}

To tackle this ill-posed correspondence subproblem, we estimate the affine parameters from the optical flow predicted by the efficient pre-trained network NeuFlowV2~\cite{zhang2025neuflow}.
Specifically, having obtained the per-pixel optical flow $\mat{F}^{k \leftarrow r} \in \mathbb{R}^{H \times W \times 2}$, we define the translation to be the displacement of the splat center, $\mat{t}_i^k = \mat{F}^{k \leftarrow r}(\bar{\mat{\mu}}_i^r)$ and estimate the linear mapping $A_i^k$ using least-squares:
Let $\mathrm{Warp}(\mat{u}) = \mat{u} + \mat{F}^{k \leftarrow r}(\mat{u})$ denote the flow warping function.
$A_i^k$ is recovered as the best candidate that maps points in  $\Omega(\bar{\mathcal{G}}_i^r)$ to $\mathrm{Warp}\left(\Omega(\bar{\mathcal{G}}_i^r)\right)$:
\begin{equation}
    A_i^k = \argmin_{A \in \mathrm{GL}(2)} \sum_{\mat{x} \in \Omega(\bar{\mathcal{G}}_i^r)} \norm{A \Delta\mat{x}^r - \Delta\mat{x}^k}_2^2,
    \label{eq:affine-correspondence}
\end{equation}
where $\Delta\mat{x}^r = \mat{x} - \bar{\mat{\mu}}_i^r$ and $\Delta\mat{x}^k = \mathrm{Warp}(\mat{x}) - \mathrm{Warp}(\bar{\mat{\mu}}_i^r)$ are the displacements of a support point $\mat{x}$ from the center in view $r$ and after warping to view $k$, respectively.

Evaluating the objective in \Cref{eq:affine-correspondence} over every pixel of $\Omega(\bar{\mathcal{G}}_i^r)$ is both wasteful and sensitive to flow noise.
Inspired by the unscented transform~\cite{uhlmann1995dynamic}, we instead constrain test points $\mat{x}$ of each splat to $2n+1 = 5$ sigma points (for $n = 2$ dimensions) placed at the center and along the principal axes:
\begin{equation}
    \mat{x}_0 = \bar{\mat{\mu}}; \;\;\; \mat{x}_{l = 1 \dots 4} = \bar{\mat{\mu}} + c_l L_l,
\end{equation}
where the spreads are $c_1 = -c_2 = c_3 = -c_4 = \sqrt{2}$, $L_1 = L_2$ is the first column of $R(\theta) \bar{S}$, and $L_3 = L_4$, the second, so that $\mat{x}_{1\dots4}$ lie one standard deviation away on either side of each axis (\Cref{fig:architecture}).
As a linear least-squares problem, \Cref{eq:affine-correspondence} under this approximation admits a closed-form solution by solving its $4 \times 4$ normal equation.
To reject unreliable correspondences, we additionally compute the reverse flow $\mat{F}^{r \leftarrow k}$ to estimate the inverse translation $\mat{t}_k^i$, and discard correspondences with $\norm{\mat{t}_k^i - \mat{t}_i^k}_2 > 3$ pixels (forward--backward consistency).
The splat tracking procedures are depicted in the middle column of \Cref{fig:architecture}.

\subsection{Geometric Splat Triangulation}\label{sec:method:triangulation}

Given the corresponding 2D splats $\{\bar{\mathcal{G}}_i^k\}_{k \in [K]}$ across all views, camera extrinsics $\{T^k\}_{k \in [K]}$ and  intrinsics $C$, we recover each 3D splat $\mathcal{G}_i$ independently by solving the well-posed projection matching problem
\begin{equation} \label{eq:g2sr-opt}
    \hat{\mathcal{G}}_i = \argmin_{\mathcal{G}_i}\, \sum_{k \in [K]}\, d_H^2\!\left(\pi^k(\mathcal{G}_i),\, \bar{\mathcal{G}}_i^k \right),
\end{equation}
instantiated from \Cref{eq:gt-proj-opt} by substituting the observed projection targets $\bar{\mathcal{G}}_i^k$ for the unobservable ground-truth projections $\pi^k(\mathcal{G}_i^*)$ and choosing the distance $d$ to be the squared Hellinger distance between two Gaussians,
\begin{equation}
    d^2_H(\mathcal{G}_1, \mathcal{G}_2) = 1 - \frac{\det(\Sigma_1 \Sigma_2)^{1/4}}{\det(\Sigma_m)^{1/2}}\exp\!\left(-\frac{\norm{\mat{\mu}_1-\mat{\mu}_2}_{\Sigma_m}^2}{8}\right),
    \label{eq:hellinger}
\end{equation}
with $\Sigma_m = \frac{1}{2}(\Sigma_1 + \Sigma_2)$.
Smooth and bounded in $[0, 1]$, $d_H^2$ jointly penalizes discrepancies in splat position, orientation, and scale within a single differentiable measure.

Since \Cref{eq:g2sr-opt} admits no closed-form solution, we minimize it iteratively with the Gauss--Newton algorithm, with all splats optimized in parallel (subscript $i$ dropped below for clarity).
At each iteration, we evaluate the per-view Jacobian
\begin{equation}
    \mat{J}^k = \pdv{d^2_H}{\pi^k(\mathcal{G})} \circ \pdv{\pi^k(\mathcal{G})}{(\mat{\mu}, \mat{\Sigma})} \circ \pdv{(\mat{\mu}, \mat{\Sigma})}{(\mat{\mu}, R, s_{u}, s_{v})}
\end{equation}
and the residual $r^k = d_H(\pi^k(\mathcal{G}),\, \bar{\mathcal{G}}^k)$, and solve for the update $\delta \mathcal{G} = (\delta\mat{\mu}, \delta R, \delta s_{u}, \delta s_{v}) = -\mat{H}^{-1} \mat{b}$, where
\begin{equation}
    \mat{H} = \sum_{k \in [K]} (\mat{J}^k)^T \mat{J}^k,\qquad\mat{b} = \sum_{k \in [K]} (\mat{J}^k)^T r^k.
\end{equation}
The update is then applied to the respective parameters as
\begin{equation}
    \mat{\mu} \leftarrow \mat{\mu} + \delta \mat{\mu},\quad
    R \leftarrow R \operatorname{Exp}(\delta R),\quad
    s_{u,v} \leftarrow s_{u,v} e^{\delta s_{u,v}}
\end{equation}
where the exponential retraction $\operatorname{Exp}$ updates $R$ on the $\mathrm{SO}(3)$ manifold and the multiplicative update keeps the scales positive.
We further scale the update by a step size $\eta$ to stabilize the optimization, and run up to 20 iterations.

Because $d_H^2$ in \Cref{eq:hellinger} is nonlinear and nonconvex, the Gauss--Newton iterations require careful initialization of each parameter group to converge to the correct splat.
We detail our initializations for mean, rotation, and scale below.

\noindent\textbf{Mean.}
The exponential term in \Cref{eq:hellinger} saturates $d_H$ to nearly 1 with a vanishing gradient when the projected and observed centers are far apart.
We overcome this narrow convergence radius by triangulating $\mat{\mu}$ using direct linear transformation (DLT)~\cite{hartley1997triangulation} triangulation from the observed 2D centers $\{\bar{\mat{\mu}}^k\}$ and the camera parameters $C$ and $\{T^k\}$.

\noindent\textbf{Rotation.}
Although corresponding splats $\{\bar{\mathcal{G}}_i^k\}_{k \in [K]}$ from only 2 views already supply $2 \times 5 = 10$ degrees-of-freedom (DoF) of geometric constraints, which is sufficient to determine the 8-DoF 3D splat, jointly recovering $R$ and $(s_u, s_v)$ from $d_H^2$ alone is ill-conditioned.
For example, an elongated but slanted splat may produce a nearly identical projection $\pi^k(\mathcal{G})$ as that of a round splat parallel to the image plane.
This effect leaves the orientation only weakly observed under projection matching.
We therefore exploit the relative affine map between view pairs, $A^{kl} = A^k (A^l)^{-1}$, obtained from \Cref{eq:affine-correspondence}, to initialize the splat's normal direction $R\mat{e}_z$ by
\begin{equation}
    R\mat{e}_z \leftarrow \argmin_{\norm{\mat{n}}_2 = 1} \norm{M \mat{n}}_2^2,
    \label{eq:triangulation:rot-init}
\end{equation}
where $M \in \mathbb{R}^{6 \times 3}$ is constructed from $A^{kl}$ and the Jacobians of the per-view projections $\pi^k$ (Eq.~(8) of~\cite{eichhardt2017computer}).
Minimizing \Cref{eq:triangulation:rot-init} seeks the normal $\mat{n}$ for which the relative affine matrix $\tilde{A}^{kl}(\mat{n})$ between view $l$ and $k$ induced by a surface orientation $\mat{n}$ best matches the measured $A^{kl}$ element-wise up to scale.
The remaining DoF, \ie, the in-plane orientation of the $\mat{e}_u$ axis about $R\mat{e}_z$, is initialized randomly and refined in the subsequent Gauss--Newton iterations.

\noindent\textbf{Scale.}
Due to the same orientation--scale ambiguity that makes splat extent $(s_u, s_v)$ difficult to recover from scratch, we initialize them by computing, for each view, the scaling that stretches $\pi^k(\mathcal{G})$ to match the extents of $\bar{\mathcal{G}}^k$ along its two axes, and take the minimum across views as a conservative estimate.

Because this initialization is derived purely from the observed projections, it adapts automatically to the metric scene scale and various camera configurations, allowing the pipeline to operate out of the box without the per-dataset fine-tuning commonly required by end-to-end methods.


\dmrecord{g2ba-unfused}{
  depth-error-inferno.room-0-2-2 = results-export/depth_estimation/rep2/images/g2ba/room_0_2_2__780__depth_error_inferno.jpg,
  depth-error-inferno.room-0-2-4 = results-export/depth_estimation/rep2/images/g2ba/room_0_2_4__658__depth_error_inferno.jpg,
  depth-turbo.room-0-2-2         = results-export/depth_estimation/rep2/images/g2ba/room_0_2_2__780__depth_turbo.jpg,
  depth-turbo.room-0-2-4         = results-export/depth_estimation/rep2/images/g2ba/room_0_2_4__658__depth_turbo.jpg,
  rep2.abs-diff                  = 0.093,
  rep2.abs-diff@fmt              = \markfst,
  rep2.abs-rel                   = 4.0,
  rep2.abs-rel@fmt               = \markfst,
  rep2.delta-125                 = 98.0,
  rep2.delta-125@fmt             = \markfst,
  rep2.fps                       = 63.2,
  rep2.fps@fmt                   = \markfst,
  rep2.lpips                     = 0.487,
  rep2.peak-inf-mb               = 88.1,
  rep2.peak-inf-mb@fmt           = \markfst,
  rep2.psnr                      = 14.01,
  rep2.ssim                      = 0.616,
  rep2.total-mem-mb              = 115.4,
  rep2.total-mem-mb@fmt          = \markfst,
  rep2.weights-mb                = 27.3,
  rep2.weights-mb@fmt            = \markfst,
  rgb.room-0-2-2                 = results-export/depth_estimation/rep2/images/g2ba/room_0_2_2__780__rgb.jpg,
  rgb.room-0-2-4                 = results-export/depth_estimation/rep2/images/g2ba/room_0_2_4__658__rgb.jpg,
}

\dmrecord{g2ba-unfused}{
  rep3.abs-diff         = 0.110,
  rep3.abs-diff@fmt     = \markfst,
  rep3.abs-rel          = 4.7,
  rep3.abs-rel@fmt      = \markfst,
  rep3.delta-125        = 97.3,
  rep3.delta-125@fmt    = \markfst,
  rep3.fps              = 28.6,
  rep3.fps@fmt          = \marksnd,
  rep3.lpips            = 0.448,
  rep3.peak-inf-mb      = 263.3,
  rep3.peak-inf-mb@fmt  = \markfst,
  rep3.psnr             = 16.19,
  rep3.ssim             = 0.704,
  rep3.total-mem-mb     = 290.6,
  rep3.total-mem-mb@fmt = \markfst,
  rep3.weights-mb       = 27.3,
  rep3.weights-mb@fmt   = \markfst,
}

\dmrecord{g2ba-unfused}{
  sca2.abs-diff         = 0.135,
  sca2.abs-diff@fmt     = \markfst,
  sca2.abs-rel          = 6.8,
  sca2.abs-rel@fmt      = \markfst,
  sca2.delta-125        = 94.5,
  sca2.delta-125@fmt    = \markfst,
  sca2.fps              = 65.3,
  sca2.fps@fmt          = \markfst,
  sca2.lpips            = 0.540,
  sca2.peak-inf-mb      = 88.1,
  sca2.peak-inf-mb@fmt  = \markfst,
  sca2.psnr             = 11.94,
  sca2.ssim             = 0.539,
  sca2.total-mem-mb     = 115.4,
  sca2.total-mem-mb@fmt = \markfst,
  sca2.weights-mb       = 27.3,
  sca2.weights-mb@fmt   = \markfst,
}

\dmrecord{g2ba-unfused}{
  sca3.abs-diff         = 0.164,
  sca3.abs-diff@fmt     = \markfst,
  sca3.abs-rel          = 7.9,
  sca3.abs-rel@fmt      = \markfst,
  sca3.delta-125        = 93.1,
  sca3.delta-125@fmt    = \markfst,
  sca3.fps              = 28.8,
  sca3.fps@fmt          = \marksnd,
  sca3.lpips            = 0.520,
  sca3.peak-inf-mb      = 263.3,
  sca3.peak-inf-mb@fmt  = \markfst,
  sca3.psnr             = 13.91,
  sca3.ssim             = 0.610,
  sca3.total-mem-mb     = 290.6,
  sca3.total-mem-mb@fmt = \markfst,
  sca3.weights-mb       = 27.3,
  sca3.weights-mb@fmt   = \markfst,
}

\dmrecord{g2ba-unfused}{
  poisson-0-2-9.accuracy            = 2.80,
  poisson-0-2-9.accuracy@fmt        = \marksnd,
  poisson-0-2-9.chamfer             = 3.03,
  poisson-0-2-9.chamfer.scan105     = 2.45,
  poisson-0-2-9.chamfer.scan105@fmt = \marksnd,
  poisson-0-2-9.chamfer.scan106     = 3.28,
  poisson-0-2-9.chamfer.scan106@fmt = \marksnd,
  poisson-0-2-9.chamfer.scan110     = 3.11,
  poisson-0-2-9.chamfer.scan110@fmt = \marksnd,
  poisson-0-2-9.chamfer.scan114     = 1.57,
  poisson-0-2-9.chamfer.scan114@fmt = \marksnd,
  poisson-0-2-9.chamfer.scan118     = 3.49,
  poisson-0-2-9.chamfer.scan122     = 3.21,
  poisson-0-2-9.chamfer.scan122@fmt = \marksnd,
  poisson-0-2-9.chamfer.scan24      = 3.41,
  poisson-0-2-9.chamfer.scan24@fmt  = \marksnd,
  poisson-0-2-9.chamfer.scan37      = 3.79,
  poisson-0-2-9.chamfer.scan37@fmt  = \marksnd,
  poisson-0-2-9.chamfer.scan40      = 3.19,
  poisson-0-2-9.chamfer.scan40@fmt  = \marksnd,
  poisson-0-2-9.chamfer.scan55      = 1.94,
  poisson-0-2-9.chamfer.scan55@fmt  = \marksnd,
  poisson-0-2-9.chamfer.scan63      = 3.46,
  poisson-0-2-9.chamfer.scan63@fmt  = \marksnd,
  poisson-0-2-9.chamfer.scan65      = 3.73,
  poisson-0-2-9.chamfer.scan65@fmt  = \marksnd,
  poisson-0-2-9.chamfer.scan69      = 2.80,
  poisson-0-2-9.chamfer.scan69@fmt  = \marksnd,
  poisson-0-2-9.chamfer.scan83      = 3.05,
  poisson-0-2-9.chamfer.scan83@fmt  = \marksnd,
  poisson-0-2-9.chamfer.scan97      = 2.91,
  poisson-0-2-9.chamfer.scan97@fmt  = \marksnd,
  poisson-0-2-9.chamfer@fmt         = \marksnd,
  poisson-0-2-9.completeness        = 3.26,
  poisson-0-2-9.peak-inf-mb         = 179.1,
  poisson-0-2-9.peak-inf-mb@fmt     = \markfst,
  poisson-0-2-9.scan-fps            = 11.78,
  poisson-0-2-9.total-mem-mb        = 206.4,
  poisson-0-2-9.total-mem-mb@fmt    = \markfst,
  poisson-0-2-9.weights-mb          = 27.3,
  poisson-0-2-9.weights-mb@fmt      = \markfst,
}

\dmrecord{g2ba-unfused}{
  tsdf-23-24-33.accuracy            = 2.92,
  tsdf-23-24-33.accuracy@fmt        = \markfst,
  tsdf-23-24-33.chamfer             = 3.39,
  tsdf-23-24-33.chamfer.scan105     = 3.41,
  tsdf-23-24-33.chamfer.scan105@fmt = \marksnd,
  tsdf-23-24-33.chamfer.scan106     = 3.98,
  tsdf-23-24-33.chamfer.scan110     = 3.83,
  tsdf-23-24-33.chamfer.scan110@fmt = \marksnd,
  tsdf-23-24-33.chamfer.scan114     = 2.05,
  tsdf-23-24-33.chamfer.scan114@fmt = \markfst,
  tsdf-23-24-33.chamfer.scan118     = 3.56,
  tsdf-23-24-33.chamfer.scan118@fmt = \markfst,
  tsdf-23-24-33.chamfer.scan122     = 2.75,
  tsdf-23-24-33.chamfer.scan122@fmt = \markfst,
  tsdf-23-24-33.chamfer.scan24      = 4.60,
  tsdf-23-24-33.chamfer.scan37      = 4.19,
  tsdf-23-24-33.chamfer.scan37@fmt  = \markfst,
  tsdf-23-24-33.chamfer.scan40      = 2.79,
  tsdf-23-24-33.chamfer.scan40@fmt  = \markfst,
  tsdf-23-24-33.chamfer.scan55      = 2.37,
  tsdf-23-24-33.chamfer.scan55@fmt  = \markfst,
  tsdf-23-24-33.chamfer.scan63      = 3.49,
  tsdf-23-24-33.chamfer.scan63@fmt  = \markfst,
  tsdf-23-24-33.chamfer.scan65      = 3.24,
  tsdf-23-24-33.chamfer.scan65@fmt  = \markfst,
  tsdf-23-24-33.chamfer.scan69      = 3.01,
  tsdf-23-24-33.chamfer.scan69@fmt  = \markfst,
  tsdf-23-24-33.chamfer.scan83      = 4.06,
  tsdf-23-24-33.chamfer.scan83@fmt  = \marksnd,
  tsdf-23-24-33.chamfer.scan97      = 3.49,
  tsdf-23-24-33.chamfer.scan97@fmt  = \marksnd,
  tsdf-23-24-33.chamfer@fmt         = \markfst,
  tsdf-23-24-33.completeness        = 3.85,
  tsdf-23-24-33.completeness@fmt    = \markfst,
  tsdf-23-24-33.peak-inf-mb         = 264.7,
  tsdf-23-24-33.peak-inf-mb@fmt     = \markfst,
  tsdf-23-24-33.scan-fps            = 21.08,
  tsdf-23-24-33.scan-fps@fmt        = \marksnd,
  tsdf-23-24-33.total-mem-mb        = 292.0,
  tsdf-23-24-33.total-mem-mb@fmt    = \markfst,
  tsdf-23-24-33.weights-mb          = 27.3,
  tsdf-23-24-33.weights-mb@fmt      = \markfst,
}

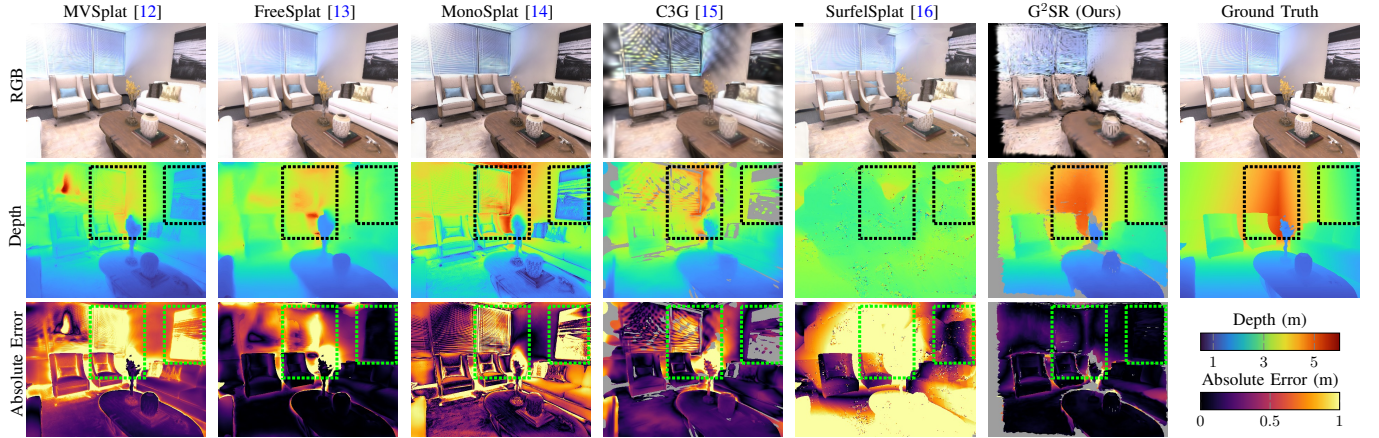
\begin{figure*}[t]
    \tikzstyle{dotted_box}=[line width=1.75pt, densely dotted]
    \newcommand{\drawdottedbox}[2]{
        \setsepchar{;/,}
        \ignoreemptyitems
        \readlist*\boxlist{#1}
        \raisebox{-0.5\height}{
            \begin{tikzpicture}
                \node (fig) [anchor=south west,inner sep=0] at (0,0) {#2};
                draw each boxes
                \foreachitem\x\in\boxlist[]{
                    \draw[dotted_box, {\boxlist[\xcnt,5]}] (\boxlist[\xcnt,1], \boxlist[\xcnt,2]) rectangle (\boxlist[\xcnt,3], \boxlist[\xcnt,4]);
                }
            \end{tikzpicture}
        }
    }

    \newcommand{\drawcolorbar}[5]{%
      \pgfplotscolorbardrawstandalone[
        colormap={example}{
            of colormap={#1},
        },
        point meta min=#2,
        point meta max=#3,
        colorbar horizontal,
        colorbar style={
          title=#5,
          title style={yshift=-1.6ex},
          width=8em,
          height=1em,
          xtick={#4},
          xticklabel style={font=\small},
        },
      ]%
    }
    
    \newcommand{\hlrects}{2.8, 2.65, 3.6, 1.5, black; 1.3, 2.65, 2.4, 1.2, black}
    \newcommand{\hlrectsgreen}{2.8, 2.65, 3.6, 1.5, green; 1.3, 2.65, 2.4, 1.2, green}

    \newcommand{\rowonekey}{rgb.room-0-2-4}
    \newcommand{\rowtwokey}{depth-turbo.room-0-2-4}
    \newcommand{\rowthreekey}{depth-error-inferno.room-0-2-4}
    
    \newcommand{\showimg}[2]{\includegraphics[width=0.2\linewidth]{\dmvp{#1}{#2}}}
    
    \centering
    \resizebox{\linewidth}{!}{
        \setlength{\tabcolsep}{-5pt}
        \begin{tabular}{cccccccc}
            &
            MVSplat~\cite{chen2024mvsplat} & FreeSplat~\cite{wang2024freesplat} & MonoSplat~\cite{liu2025monosplat} & C3G~\cite{an2025c3g} & SurfelSplat~\cite{daisurfelsplat} & G$^2$SR~(Ours) & Ground Truth \\

            \rotatebox[origin=c]{90}{RGB} &
            \drawdottedbox{;}{\showimg{mvsplat}{\rowonekey}} &
            \drawdottedbox{;}{\showimg{freesplat}{\rowonekey}} &
            \drawdottedbox{;}{\showimg{monosplat}{\rowonekey}} &
            \drawdottedbox{;}{\showimg{c3g}{\rowonekey}} &
            \drawdottedbox{;}{\showimg{surfelsplat}{\rowonekey}} &
            \drawdottedbox{;}{\showimg{g2ba}{\rowonekey}} &
            \drawdottedbox{;}{\showimg{gt}{\rowonekey}}
            
            \\[3.8em]

            \rotatebox[origin=c]{90}{Depth} &
            \drawdottedbox{\hlrects}{\showimg{mvsplat}{\rowtwokey}} &
            \drawdottedbox{\hlrects}{\showimg{freesplat}{\rowtwokey}} &
            \drawdottedbox{\hlrects}{\showimg{monosplat}{\rowtwokey}} &
            \drawdottedbox{\hlrects}{\showimg{c3g}{\rowtwokey}} &
            \drawdottedbox{\hlrects}{\showimg{surfelsplat}{\rowtwokey}} &
            \drawdottedbox{\hlrects}{\showimg{g2ba}{\rowtwokey}} &
            \drawdottedbox{\hlrects}{\showimg{gt}{\rowtwokey}}

            \\[3.8em]

            \rotatebox[origin=c]{90}{Absolute Error} &
            \drawdottedbox{\hlrectsgreen}{\showimg{mvsplat}{\rowthreekey}} &
            \drawdottedbox{\hlrectsgreen}{\showimg{freesplat}{\rowthreekey}} &
            \drawdottedbox{\hlrectsgreen}{\showimg{monosplat}{\rowthreekey}} &
            \drawdottedbox{\hlrectsgreen}{\showimg{c3g}{\rowthreekey}} &
            \drawdottedbox{\hlrectsgreen}{\showimg{surfelsplat}{\rowthreekey}} &
            \drawdottedbox{\hlrectsgreen}{\showimg{g2ba}{\rowthreekey}} &

            \raisebox{-0.5\height}{
                \begin{tikzpicture}
                    \begin{scope}[yshift=1.2cm]
                        \drawcolorbar{turbo}{0.5}{6}{1, 3, 5}{Depth (m)}
                    \end{scope}
                    \begin{scope}[yshift=0cm]
                        \drawcolorbar{inferno}{0}{1}{0, 0.5, 1}{Absolute Error (m)}
                    \end{scope}
                \end{tikzpicture}
            }

            \\

        \end{tabular}
    }
    \caption{Qualitative comparison on a Replica 2-view reconstruction. Compared with baselines, G$^2$SR's rendered depth (Row 2) best approaches ground truth with uniformly low error (Row 3), especially in areas with fine texture (right rectangle) and at large depths (central rectangle). Although not optimized for, overall scene appearance is recovered despite the artifacts introduced by gaps between splats or non-covisible regions around image boundaries (Row 1).}
    \label{fig:visualizations}
\end{figure*}

\begin{table*}[!t]
    \centering
    \caption{Depth estimation accuracy and efficiency on the Replica~\cite{straub2019replica} and ScanNet~\cite{dai2017scannet} datasets.}
    \label{tab:depth-est-replica-scannet}
    \setlength{\tabcolsep}{4pt}
    
    \newcommand{\metrics}[2]{
        \dmvf{#1}{rep#2.abs-diff}     / \dmvf{#1}{sca#2.abs-diff} &
        \dmvf{#1}{rep#2.abs-rel}      / \dmvf{#1}{sca#2.abs-rel} &
        \dmvf{#1}{rep#2.delta-125}    / \dmvf{#1}{sca#2.delta-125} &
        \dmvf{#1}{rep#2.total-mem-mb} / \dmvf{#1}{sca#2.total-mem-mb} &
        \dmvf{#1}{rep#2.fps}          / \dmvf{#1}{sca#2.fps}
    }

    \resizebox{\textwidth}{!}{
    \begin{threeparttable}
        \begin{tabular}{@{}lccccccccccc}
            \toprule
            \multirow{2}{*}{Method} & \multicolumn{5}{c}{2-view (Replica / ScanNet)} & \multicolumn{5}{c}{3-view (Replica / ScanNet)} \\
            \cmidrule(lr){2-6} \cmidrule(lr){7-11}  & \multicolumn{1}{c}{Abs Diff (m) $\downarrow$} & \multicolumn{1}{c}{Abs Rel (\%) $\downarrow$} & \multicolumn{1}{c}{$\delta_1 (\%) \uparrow$} & \multicolumn{1}{c}{Memory (MB) $\downarrow$} & \multicolumn{1}{c}{RPS\tnote{1} $\uparrow$} & \multicolumn{1}{c}{Abs Diff (m) $\downarrow$} & \multicolumn{1}{c}{Abs Rel (\%) $\downarrow$} & \multicolumn{1}{c}{$\delta_1 (\%) \uparrow$} & \multicolumn{1}{c}{Memory (MB) $\downarrow$} & \multicolumn{1}{c}{RPS\tnote{1} $\uparrow$} \\ \midrule
            
            MVSplat~\cite{chen2024mvsplat}     & \metrics{mvsplat}{2}     & \metrics{mvsplat}{3}     \\
            FreeSplat~\cite{wang2024freesplat} & \metrics{freesplat}{2}   & \metrics{freesplat}{3}   \\
            MonoSplat~\cite{liu2025monosplat}  & \metrics{monosplat}{2}   & \metrics{monosplat}{3}   \\
            C3G~\cite{an2025c3g}               & \metrics{c3g}{2}         & \metrics{c3g}{3}         \\
            SurfelSplat~\cite{daisurfelsplat}  & \metrics{surfelsplat}{2} & \metrics{surfelsplat}{3} \\
            G$^2$SR (Ours)                     & \metrics{g2ba}{2}        & \metrics{g2ba}{3}        \\
            
            \bottomrule
        \end{tabular}

        \begin{tablenotes}
            \item All methods \textit{except} G$^2$SR are median-aligned to metric scale using ground truth depth before evaluating.
            \item[1] RPS: Reconstructions per second. Rate of splat reconstruction from a set of 2 or 3 views.
        \end{tablenotes}
    \end{threeparttable}
    }
\end{table*}

\begin{table}[!t]
    \centering
    
    \caption{\scriptsize Mesh reconstruction accuracy and efficiency on the DTU~\cite{jensen2014dtu} dataset.}
    \label{tab:dtu-mesh}
    
    \dmrecord{c3g}{
      tsdf-23-24-33.chamfer = -,
      tsdf-23-24-33.total-mem-mb = \dmvp{c3g}{poisson-0-2-9.total-mem-mb},
      tsdf-23-24-33.scan-fps = \dmvp{c3g}{poisson-0-2-9.scan-fps}
    }
    
    \newcommand{\metrics}[1]{\dmvf{#1}{tsdf-23-24-33.chamfer} & \dmvf{#1}{poisson-0-2-9.chamfer} & \dmvf{#1}{tsdf-23-24-33.total-mem-mb} & \dmvf{#1}{tsdf-23-24-33.scan-fps}}

    \resizebox{0.48\textwidth}{!}{%
    
    \begin{threeparttable}
        \begin{tabular}{lrrrrrrrr}
            \toprule
            Method & CD (T)\tnote{1} $\downarrow$ & CD (P)\tnote{2} $\downarrow$ & Memory (MB) $\downarrow$ & RPS $\uparrow$ \\
            \midrule
            
            MVSplat~\cite{chen2024mvsplat} & \metrics{mvsplat} \\
            FreeSplat~\cite{wang2024freesplat} & \metrics{freesplat} \\
            MonoSplat~\cite{liu2025monosplat} & \metrics{monosplat} \\
            C3G~\cite{an2025c3g} & \metrics{c3g} \\
            SurfelSplat~\cite{daisurfelsplat} & \metrics{surfelsplat} \\
            G$^2$SR (Ours) & \metrics{g2ba} \\
            \bottomrule
        \end{tabular}
        
        \begin{tablenotes}
            \item[1, 2] CD (T) / (P): Mean chamfer distance in mm between ground truth and mesh reconstructed with TSDF fusion or Poisson reconstruction.
        \end{tablenotes}
    \end{threeparttable}
    
    }%
\end{table}

\section{Experiments}~\label{sec:exp}

We test G$^2$SR against state-of-the-art (SOTA) end-to-end 3DGS methods: \textit{MVSplat}~\cite{chen2024mvsplat}, \textit{FreeSplat}~\cite{wang2024freesplat}, \textit{MonoSplat}~\cite{liu2025monosplat}, \textit{C3G}~\cite{an2025c3g}, and \textit{SurfelSplat}~\cite{daisurfelsplat}, on efficiency (memory and throughput) and geometric accuracy (depth and mesh error).
Across 2- and 3-view reconstruction tasks on three datasets, G$^2$SR leads in geometric accuracy in almost all experiments while performing \textbf{\tpmin--\tpmax~reconstructions per second (RPS)} on a single NVIDIA RTX 4090 GPU at $384 \times 512$ resolution and only consumes \textbf{\memmin--\memmax~MB} of GPU memory.
Compared to baselines, G$^2$SR achieves a speedup of up to \textbf{\speedupxmax$\times$} and memory reduction of \textbf{\memredxmin--\memredxmax$\times$}.
In the remainder of this section, \Cref{sec:exp:exp-setup} describes the experimental setup; \Cref{sec:exp:efficiency} quantifies the efficiency of each method, and \Cref{sec:exp:geom-acc} reports the accuracy.
For completeness, we report view-synthesis metrics in \Cref{sec:exp:rend-acc}, where G$^2$SR stays competitive in per-pixel and structural fidelity (PSNR, SSIM) while trailing in fine perceptual texture (LPIPS) (\Cref{fig:visualizations}).

\subsection{Experiment Setup}~\label{sec:exp:exp-setup}

\subsubsection{Datasets}

We benchmark on three datasets spanning complementary conditions.
\textbf{Replica}~\cite{straub2019replica} provides high-fidelity synthetic indoor renders with noiseless RGB and dense ground-truth depth for assessing the theoretical upper bound of reconstruction accuracy.
\textbf{ScanNet}~\cite{dai2017scannet} captures real-world indoor scans with realistic image noise, lighting changes, and motion blur for probing robustness.
\textbf{DTU}~\cite{jensen2014dtu} provides object-centric captures with mesh ground truth for evaluating surface reconstruction quality under controlled lighting.
For Replica and ScanNet, we adopt the splits processed by FreeSplat~\cite{wang2024freesplat}, which provide pre-sampled 2- and 3-view sets with rectified intrinsics and field of view.
For DTU, we adopt the 2DGS~\cite{wu2025sparse2dgs} variant, which normalizes each scene to a unit bounding cube.
%

\subsubsection{Tasks and Metrics}

We evaluate two reconstruction tasks: \textit{depth estimation} on Replica and ScanNet, and \textit{mesh reconstruction} on DTU.
In depth estimation, each method is given a set of 2 or 3 views designated by FreeSplat~\cite{wang2024freesplat}'s evaluation protocol as context to predict the Gaussian splats representing the scene; the splats are then rendered into depth maps at another set of target views and compared against the ground truth depth maps.
We report Absolute Difference (\textit{Abs Diff}, m), Absolute Relative Error (\textit{Abs Rel}, \%), and the inlier ratio within $\pm 25\%$ of ground truth ($\delta_1$, \%) on target views.
For mesh reconstruction, we evaluate all 15 DTU scans and report the mean Chamfer Distance (\textit{CD}, mm) between the ground-truth and reconstructed meshes.
We construct the mesh in two ways: \textit{TSDF fusion} on rendered depth, following NeuS~\cite{wang2021neus} with context views $\{23, 24, 33\}$; and \textit{Poisson reconstruction} on splat centers, following SurfelSplat~\cite{daisurfelsplat} with context views $\{0, 2, 9\}$.
While reporting both for completeness, we note that the more widely adopted TSDF protocol renders intermediate depth maps from the splats and therefore holistically tests scale and rotation with the center, whereas Poisson isolates the accuracy of splat centers.
We adopt a resolution of $384 \times 512$ for depth estimation and $480 \times 640$ for mesh reconstruction.

In addition to accuracy metrics, we report two quantities to characterize the \textit{efficiency}:
\textit{Memory} (MB) is the peak size of GPU memory allocated during a single 2- or 3-view reconstruction, comprising model weights, activations, and other intermediate buffers, as recorded via PyTorch's memory allocation trace.
\textit{Reconstructions per second} (RPS) measures the rate of the same.
For DTU mesh reconstruction, meshing time is excluded so that RPS reflects splat inference alone.

We additionally report view-synthesis metrics, including PSNR, SSIM, and LPIPS for completeness, although G$^2$SR is not optimized to produce color-accurate splats that capture fine-grained textural details as baselines do.
All methods are evaluated over valid pixels with rendered opacity $\alpha \geq 0.5$ and report the resulting \textit{coverage} (the percentage of such pixels), so that unreconstructed regions are not counted against a method.

\subsubsection{Baselines}

We compare against five SOTA end-to-end few-view reconstruction methods spanning the common backbone choices:
\textit{MVSplat}~\cite{chen2024mvsplat} (Custom CNN + Swin attention, hybrid), 
\textit{FreeSplat}~\cite{wang2024freesplat} (EfficientNet, CNN),
\textit{MonoSplat}~\cite{liu2025monosplat} (DepthAnythingV2-S, ViT), 
\textit{C3G}~\cite{an2025c3g} (VGGT-1B, ViT), 
and \textit{SurfelSplat}~\cite{daisurfelsplat} (DINOv2-L, ViT). 
All baselines are evaluated using their released checkpoints and default hyperparameters.

\subsubsection{Implementation Details}~\label{sec:exp:exp-setup:impl-details}
We only train the splat detection network $\Phi$ (\Cref{sec:method:detection}) on RealEstate10K~\cite{zhou2018stereo} and use the pretrained NeuFlowV2~\cite {zhang2025neuflow} optical flow network off-the-shelf.
No fine-tuning is performed on Replica, ScanNet, or DTU.
The Gauss-Newton solver runs for $N=20$ iterations.

When evaluating depth estimation, since the baselines do not always predict splats in metric scale, we median-align their predictions to the ground truth before computing depth metrics to make a meaningful comparison\footnote{C3G internally normalizes the camera baseline to unit length, while the others inherit the depth scale from their training data. However, note that the ground truth is not available during online reconstruction.}.
In contrast, G$^2$SR produces metric-scale depth by construction through triangulation and is evaluated \emph{unaligned}.
Every depth result reported below thus reflects a fairness advantage granted to the baselines.
%
%
All evaluations are conducted on a desktop workstation with a single NVIDIA RTX~4090 GPU and a 24-core Intel i9-14900K CPU.
This platform serves as a controlled measurement bench on which every method, including those requiring high memory and compute, runs unconstrained.
We expect the accuracy, memory footprint, and relative throughput to transfer across platforms and defer mobile benchmarking to future work.

\subsection{Memory and Throughput}~\label{sec:exp:efficiency}

\subsubsection{Memory}~\label{sec:exp:efficiency:memory}
\Cref{tab:depth-est-replica-scannet,tab:dtu-mesh} report the peak GPU memory allocated during inference, comprising model weights, activations, and intermediate buffers.
G$^2$SR's frontend, consisting of the 0.41~M-parameter detector and NeuFlowV2, accounts for 27~MB of weights; activations peak at $\sim$\dmeval[0]{\dmv{g2ba}{rep2.peak-inf-mb}}~MB on 2-view inputs and $\sim$\dmeval[0]{\dmv{g2ba}{tsdf-23-24-33.peak-inf-mb}}~MB on 3-view inputs, for a total of \textbf{\memmin--\memmax~MB} per reconstruction under an image resolution of $384 \times 512$.
On the same workloads, the baselines consume a memory of \dmv{monosplat}{rep2.total-mem-mb}~MB to \dmeval[1]{\dmv{surfelsplat}{rep2.total-mem-mb} / 1024.0}~GB, corresponding to a memory reduction of \textbf{\memredxmin--\memredxmax$\times$} across the experiments.
By offloading well-posed structural inference to the geometric backend, our frontend is significantly smaller than the baselines', whose weights alone consume \dmeval[0]{\dmv{monosplat}{rep2.weights-mb}}~MB (MonoSplat) -- \dmeval[1]{\dmv{c3g}{rep2.weights-mb} / 1024.0}~GB (C3G).

\subsubsection{Throughput}~\label{sec:exp:efficiency:throughput}
G$^2$SR runs at \textbf{\dmv{g2ba}{sca3.fps}--\dmv{g2ba}{rep2.fps}~RPS} on depth estimation, with the lower end corresponding to 3-view inputs, and at \textbf{\dmv{g2ba}{tsdf-23-24-33.scan-fps}~RPS} on DTU mesh reconstruction.
In comparison, the baselines achieve \dmv{surfelsplat}{rep2.fps}--\dmv{monosplat}{rep2.fps}~RPS and \dmv{surfelsplat}{tsdf-23-24-33.scan-fps}--\dmv{monosplat}{tsdf-23-24-33.scan-fps}~RPS on the two tasks, respectively.
The \textbf{\speedupxmin--\speedupxmax$\times$} speedup arises from the same architectural shift that yields the memory savings: the geometric backend both allows for a simpler neural frontend and performs structural inference analytically that may require multiple neural layers to approximate.

\subsection{Geometric Accuracy}~\label{sec:exp:geom-acc}

\subsubsection{Depth Estimation}~\label{sec:exp:geom-acc:depth-estimation}
\Cref{tab:depth-est-replica-scannet} reports depth accuracy on Replica and ScanNet under 2- and 3-view inputs.
On Replica 2-view, G$^2$SR attains an absolute difference of \dmv{g2ba}{rep2.abs-diff}~m, relative error of \textbf{\dmv{g2ba}{rep2.abs-rel}\%}, and $\delta_1$ of \textbf{\dmv{g2ba}{rep2.delta-125}\%}, which roughly halves the error of the strongest baseline (FreeSplat) and retains a similar margin across the other three settings.
Notably, G$^2$SR does so without the median alignment granted to the baselines (\Cref{sec:exp:exp-setup:impl-details}), confirming that it produces metric-scale depth consistent across views.
\Cref{fig:visualizations} shows a representative Replica 2-view reconstruction: aside from boundary pixels lacking covisibility in context views and minor gaps between splats, G$^2$SR recovers most of the scene at uniformly low error, whereas end-to-end methods struggle on complex texture (\eg, the art painting on the wall) and at large depths (\eg, the back of the room).
These failure patterns reflect where the depth comes from: a network that regresses depth from learned priors is biased toward its training distribution, compressing unfamiliar depth ranges and letting image texture bleed into geometry, whereas G$^2$SR solves depth from the projective constraints, so its residual error stems from 2D correspondence noise rather than from a learned 3D bias.

\subsubsection{Mesh Reconstruction}~\label{sec:exp:geom-acc:surface}
Under the widely adopted NeuS protocol (TSDF fusion), G$^2$SR achieves the lowest Chamfer distance of \textbf{\dmv{g2ba}{tsdf-23-24-33.chamfer}~mm}, ahead of FreeSplat (\dmv{freesplat}{tsdf-23-24-33.chamfer}), MonoSplat (\dmv{monosplat}{tsdf-23-24-33.chamfer}), and the rest; under the SurfelSplat protocol (Poisson reconstruction on splat centers), G$^2$SR (\dmv{g2ba}{poisson-0-2-9.chamfer}~mm) is second only to SurfelSplat (\dmv{surfelsplat}{poisson-0-2-9.chamfer}~mm).
We note that TSDF is the more demanding protocol, as it renders intermediate depth from the splats and thus tests scale and rotation alongside the center, requiring all splat parameters to be jointly consistent.
Most baselines degrade noticeably from Poisson to TSDF, whereas G$^2$SR stays nearly unchanged, indicating that G$^2$SR recovers the full splat geometry, instead of centers only.

\subsubsection{Robustness to Camera Geometry}~\label{sec:exp:geom-acc:generalization}
Solving the well-posed 3D triangulation analytically also makes G$^2$SR robust to shifts in camera geometry without diversified training or finetuning.
The baselines fail under two shifts:
\textit{(i) Intrinsics}---methods trained on low-focal-length RealEstate10K-style data degrade (MVSplat, MonoSplat) or yield no meaningful CD (C3G) on DTU's $\sim2.5\times$ longer focal length; SurfelSplat, conversely, is trained on DTU but fails on indoor scenes (\Cref{tab:depth-est-replica-scannet,fig:visualizations}).
\textit{(ii) View configuration}---even within DTU, SurfelSplat degrades by $\sim5\times$ across protocols (\dmv{surfelsplat}{poisson-0-2-9.chamfer} to \dmv{surfelsplat}{tsdf-23-24-33.chamfer}~mm) as the test views $\{23, 24, 33\}$ differ from its training views $\{0, 2, 9\}$.
In comparison, trained only on RealEstate10K with NeuFlowV2 off-the-shelf, G$^2$SR is unaffected in both cases as the backend resolves camera parameters automatically and leaves the frontend focused on camera-agnostic 2D tasks.

\begin{table}[!t]
    \centering
    
    \caption{Rendering accuracy of 3-view reconstruction on Replica~\cite{straub2019replica} and ScanNet~\cite{dai2017scannet}. All columns reported as Replica / ScanNet.}
    \label{tab:rend-acc}
    
    \setlength{\tabcolsep}{4pt}

    \newcommand{\metrics}[1]{
        \dmv{#1}{rep3.psnr} / \dmv{#1}{sca3.psnr} &
        \dmv{#1}{rep3.ssim} / \dmv{#1}{sca3.ssim} &
        \dmv{#1}{rep3.lpips} / \dmv{#1}{sca3.lpips} &
        \dmv{#1}{rep3.coverage} / \dmv{#1}{sca3.coverage}}

    \begin{threeparttable}
        \begin{tabular}{@{}lcccc@{}}
            \toprule
            Method & PSNR $\uparrow$ & SSIM $\uparrow$ & LPIPS $\downarrow$ & Cov. (\%)\tnote{1} $\uparrow$ \\
    
            \midrule
    
            MVSplat~\cite{chen2024mvsplat}     & \metrics{mvsplat}     \\
            FreeSplat~\cite{wang2024freesplat} & \metrics{freesplat}   \\
            MonoSplat~\cite{liu2025monosplat}  & \metrics{monosplat}   \\
            C3G~\cite{an2025c3g}               & \metrics{c3g}         \\
            SurfelSplat~\cite{daisurfelsplat}  & \metrics{surfelsplat} \\
            G$^2$SR (Ours) & \metrics{g2ba} \\
            
            \bottomrule
        \end{tabular}

        \begin{tablenotes}
            \item[1] Cov.: Coverage, \ie, the percentage of valid pixels (rendered opacity $\alpha \geq 0.5$); rendering metrics are computed over valid pixels only.
        \end{tablenotes}
    \end{threeparttable}
\end{table}

\subsection{Rendering Accuracy}~\label{sec:exp:rend-acc}

\newcommand{\dbdiffrange}{\dmeval[0]{\dmv{freesplat}{sca3.psnr} - \dmv{g2ba}{sca3.psnr}}--\dmeval[0]{\dmv{freesplat}{rep3.psnr} - \dmv{g2ba}{rep3.psnr}}}

For completeness, \Cref{tab:rend-acc} reports PSNR, SSIM, and LPIPS for 3-view reconstruction on Replica and ScanNet, evaluated over each method's valid pixels.
G$^2$SR reaches a PSNR of \dmv{g2ba}{rep3.psnr} (Replica) and \dmv{g2ba}{sca3.psnr} (ScanNet), which is \dbdiffrange~dB behind the SOTA FreeSplat but still within $\sim$\dmeval[1]{\dmv{mvsplat}{rep3.psnr} - \dmv{g2ba}{rep3.psnr}}~dB of the popular MVSplat, confirming that it recovers appearances at a high level.
It trails the leading baselines on fine texture (LPIPS) but still outperforms C3G and SurfelSplat.
This gap follows from the different supervision target: while baselines learn to minimize rendering loss directly against ground-truth pixels, our formulation in \Cref{eq:g2sr-opt} fits splats to detected 2D targets that themselves only approximate the images and thus does not guarantee the transfer of fine-grained texture.

G$^2$SR's lower coverage (\dmv{g2ba}{rep3.coverage}\% / \dmv{g2ba}{sca3.coverage}\% versus above 94\% for most baselines) stems from the aggressive outlier rejection, which discards splats with inconsistent forward--backward or out-of-bounds warps (\Cref{sec:method:correspondence}).
While leaving boundary and occluded regions unreconstructed (grey in \Cref{fig:visualizations}) that an end-to-end method would inpaint, it safeguards geometric accuracy by avoiding hallucinated structure; the missing coverage can be recovered by accumulating over more views.

\section{Conclusion}
We presented G$^2$SR, an efficient framework for Gaussian-based few-view surface reconstruction that decouples the pipeline into a lightweight neural frontend for detecting and tracking 2D Gaussian splats across views, and an analytic geometric backend that triangulates them into metric-scale 3D splats.
By reserving learning for the ill-posed image-plane tasks and solving the well-posed 3D structure analytically, G$^2$SR keeps its networks small and focused on camera-agnostic 2D reasoning.
On Replica, ScanNet, and DTU, G$^2$SR matches or surpasses the depth estimation and mesh reconstruction accuracy of state-of-the-art end-to-end methods while running at \tpmin--\tpmax~RPS with only \memmin--\memmax~MB of GPU memory on an RTX 4090 GPU (2- and 3-view, resolution $> 384 \times 512$), which is up to \memredxmax$\times$ less than prior methods and makes it a practical step toward online, high-fidelity Gaussian-based surface reconstruction on resource-constrained platforms.

{
\bibliographystyle{IEEEtran}
\bibliography{references}
}

\end{document}